\documentclass{article}

\usepackage{microtype}
\usepackage{graphicx}
\usepackage{subfigure}
\usepackage{booktabs} 
\usepackage{colortbl}
\usepackage{multirow}
\usepackage{bbm}
\usepackage{enumitem}
\setenumerate[1]{itemsep=0pt, partopsep=0pt, parsep=\parskip, topsep=5pt}
\setitemize[1]{itemsep=0pt, partopsep=0pt, parsep=\parskip, topsep=5pt}
\setdescription[1]{itemsep=0pt, partopsep=0pt, parsep=\parskip, topsep=5pt}
\usepackage{hyperref}

\usepackage[accepted]{icml2024}

\usepackage{amsmath}
\usepackage{amssymb}
\usepackage{mathtools}
\usepackage{amsthm}
\usepackage{bm}
\usepackage{sidecap}
\usepackage[most]{tcolorbox}  

\usepackage[capitalize,noabbrev]{cleveref}

\usepackage{subcaption}
\usepackage{subfloat}

\definecolor{darkgrey}{rgb}{0.53,0.53,0.53}
\definecolor{mygrey}{rgb}{0.9,0.9,0.9}
\definecolor{purple}{RGB}{230, 227, 254}
\definecolor{lightgreen}{RGB}{238, 252, 241}
\definecolor{lightred}{RGB}{231, 187, 187}
\definecolor{darkred}{RGB}{198, 129, 129}

\definecolor{tabhighlight}{HTML}{e5e5e5}
\definecolor{someorange}{rgb}{0.773,0.353,0.067}
\definecolor{someblue}{rgb}{0.27, 0.35, 0.760}

\let\oldcite\cite 
\renewcommand{\cite}[1]{\textcolor{someorange}{\oldcite{#1}}}

\theoremstyle{plain}
\newtheorem{theorem}{Theorem}[section]

\theoremstyle{definition}
\newtheorem{definition}[theorem]{Definition}

\theoremstyle{remark}

\usepackage[textsize=tiny]{todonotes}

\icmltitlerunning{Model Tailor: Mitigating Catastrophic Forgetting in MLLMs}

\begin{document}

\twocolumn[
\icmltitle{Model Tailor: Mitigating Catastrophic Forgetting in \\
Multi-modal Large Language Models}



\icmlsetsymbol{equal}{*}

\begin{icmlauthorlist}
\icmlauthor{Didi Zhu}{equal,yyy,comp}
\icmlauthor{Zhongyi Sun}{comp}
\icmlauthor{Zexi Li}{yyy}
\icmlauthor{Tao Shen}{yyy}
\icmlauthor{Ke Yan}{comp}
\icmlauthor{Shouhong Ding}{comp}
\icmlauthor{Kun Kuang}{yyy}
\icmlauthor{Chao Wu}{yyy}
\end{icmlauthorlist}

\icmlaffiliation{yyy}{Department of Computer and Science, Zhejiang University}
\icmlaffiliation{comp}{Tencent Youtu Lab}

\icmlcorrespondingauthor{Kun Kuang}{kunkuang@zju.edu.cn}
\icmlcorrespondingauthor{Chao Wu}{chao.wu@zju.edu.cn}

\icmlkeywords{Machine Learning, ICML}

\vskip 0.3in
]



\printAffiliationsAndNotice{\icmlEqualContribution} 

\begin{abstract}
Catastrophic forgetting emerges as a critical challenge when fine-tuning multi-modal large language models (MLLMs), where improving performance on unseen tasks often leads to a significant performance drop on the original tasks.
This paper presents a comprehensive analysis of catastrophic forgetting in MLLMs and introduces a post-training adjustment method called Model Tailor. 
Our method primarily preserves the pre-trained parameters while replacing a small number ($\leq$ 10\%) of fine-tuned parameters, maintaining $\sim$ 99\% effectiveness on original tasks versus pre-training, and achieving $\sim$ 97\% on new tasks compared to standard fine-tuning.
Specifically, we derive a sparse mask to identify the \textit{``model patch"}, based on a fusion strategy that integrates salience and sensitivity analysis. Subsequently, a compensation mechanism is introduced to \textit{``decorate the patch"}, enhancing the model's performance on both target and original tasks. Additionally, our method is adaptable to multi-task scenarios.
Through extensive experiments on InstructBLIP and LLaVA-1.5 in both image captioning and visual question answering tasks, our approach demonstrates significant task adaptability while preserving inherent pre-trained capabilities.
\end{abstract}

\section{Introduction}

Recent advancements in large language models (LLMs) \cite{devlin2018bert, brown2020language, touvron2023llama, wu2023brief} have marked a significant milestone in the pursuit of human-like artificial intelligence.  This evolution has been accelerated by integrating additional modalities, particularly vision, leading to the emergence of multi-modal LLMs (MLLMs) \cite{li2023blip, dai2023instructblip, liu2023visual, zhu2023minigpt, achiam2023gpt}. Prominent examples such as GPT-4V~\cite{achiam2023gpt} have played a pivotal role in enriching multi-modal dialog systems, demonstrating the versatility and depth of these advanced models. 
MLLMs typically comprise modality-specific sub-modules (such as a vision encoder and an LLM) to capture knowledge from different modalities, necessitating the incorporation of a lightweight projector for achieving cross-modal alignment \cite{li2023blip, liu2023visual, zhu2023minigpt}.

\begin{figure*}[t]
  \centering
    \includegraphics[width=6.8in]{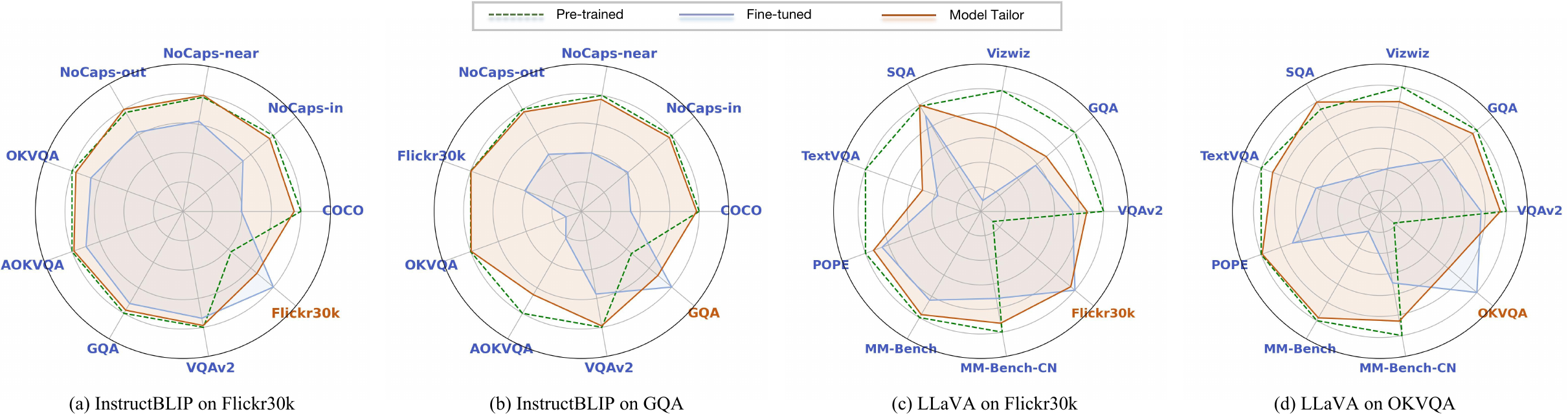}
  \caption{\textbf{Catastrophic Forgetting in Multi-modal Large Language Models.} After fine-tuning on two distinct tasks (in \textcolor{someorange}{orange}), InstructBLIP and LLaVa1.5 exhibit a significant performance decline on their original tasks (in \textcolor{someblue}{blue}). Our method offers a remedy to this issue, mitigating the adverse effects of catastrophic forgetting.
  }
  \vspace{-5mm}
  \label{fig:fig1_1}
\end{figure*}

 Due to the architectural complexity of MLLMs and the inherent disparities in data characteristics across modalities, these models face challenges in task generalization compared to uni-modal LLMs \cite{sung2023ecoflap}.
As a result, MLLMs often demonstrate notably poor performance when faced with unseen tasks \cite{lin2023speciality, zhai2023investigating, he2023continual}.
While fine-tuning MLLMs on these unseen data appears to be a straightforward solution, our empirical findings indicate that such approaches significantly impair the models' performance on their original tasks, as evidenced in \autoref{fig:fig1_1}. This observed decline, which is defined as catastrophic forgetting in MLLMs, highlights a critical challenge that has been seldom explored. This naturally raises a pivotal question: 

 \begin{tcolorbox}[notitle, rounded corners, colframe=darkgrey, colback=white, boxrule=2pt, boxsep=0pt, left=0.15cm, right=0.17cm, enhanced, shadow={2.5pt}{-2.5pt}{0pt}{opacity=5,mygrey},toprule=2pt, before skip=0.65em, after skip=0.75em 
  ]
\emph{
  {
    \centering 
  {
    \fontsize{8pt}{13.2pt}\selectfont 
    How to enhance the performance of MLLMs on target tasks without diminishing their effectiveness in original tasks? 
  }
  \\
  }
  }
\end{tcolorbox}

Current approaches to mitigate catastrophic forgetting in machine learning \cite{goodfellow2013empirical, masana2022class, yang2023neural} are primarily devised for small models and rely heavily on full-model fine-tuning.
In MLLMs, full-model fine-tuning on new tasks leads to heightened computational demands, escalating data storage and training costs, thus becoming impractical with more downstream tasks and highlighting its inefficiency for large-scale use.
Furthermore, while parameter-efficient methods like Low-Rank Adaptation (LoRA) \cite{hu2021lora} are designed to reduce computational and memory burdens, they fall short in addressing catastrophic forgetting in MLLMs. During fine-tuning with LoRA, a substantial number of redundant parameter modifications may still be retained, which ineffectively serve the target task and simultaneously erode the pre-trained knowledge (empirically verified in \autoref{fig:lora_llava}).

In response to these challenges, we introduce Model Tailor—a parameter-efficient post-training method that meticulously integrates fine-tuned parameters into the pre-trained model fabric. Just as a tailor selects patches to enhance a garment, Model Tailor discerns and modifies a minimal set of parameters, known as the \textit{``model patch"}, from the fine-tuned model to fortify the model's capabilities on new tasks without forfeiting its pre-trained proficiencies. Guided by the principles of the Lottery Ticket Hypothesis~\cite{frankle2019lottery,li2023otter},
the model patch is identified via a fusion strategy that analyzes both parameter changes and loss variations. Moreover, drawing inspiration from the Optimal Brain Surgeon (OBS) theory~\cite{lecun1989optimal, hassibi1993optimal}, we reinforce the model's grasp on target task knowledge through \textit{``patch decoration"} – a process of precise weight compensation based on the inverse of the Hessian matrix. 
As illustrated in \autoref{fig:fig1_1}, Model Tailor maximizes the restoration of the model's performance on original tasks while also enhancing its efficacy on target tasks.
Additionally, our approach demonstrates remarkable flexibility in multi-task scenarios. By stitching these distinct patches, the model can absorb task-specific knowledge while also enhancing its performance on the original task.
In summary, our work makes several significant contributions:

\vspace{-1.5mm}
\begin{itemize}[leftmargin=*]
\setlength{\parskip}{3pt}
    \item \textbf{Revealing Catastrophic Forgetting in MLLMs.} We present the comprehensive analysis of catastrophic forgetting in MLLMs, specifically in tasks like image captioning and visual question answering (VQA), highlighting challenges in multimodal generation and comprehension.   
    \item \textbf{Pioneering Parameter-Efficient Solution in MLLMs.} To the best of our knowledge, Model Tailor is the first parameter-efficient method to address catastrophic forgetting in MLLMs. By implementing the ``model patch" and ``patch decorator", our approach effectively tackles this issue and demonstrates adaptability in multi-task scenarios.
    \item \textbf{Theoretical Analysis and Empirical Validation.} Through rigorous validation, we verify the effectiveness of our proposed method. Our results show significant improvements in retaining performance on original tasks while efficiently adapting to new tasks, thereby confirming the practicality and robustness of our approach.
\end{itemize}

\section{Reltaed Work}
\label{sec:related}
\textbf{Multi-modal Large Language Models.} 
MLLMs mark a significant leap in vision-language modeling, substantially enhancing reasoning and understanding capabilities. Designed to process and interpret information across modalities, MLLMs adeptly handle complex tasks requiring deep contextual understanding. Recent advancements~\cite{li2023blip,dai2023instructblip,liu2023visual,liu2023improved,zhu2023minigpt,alayrac2022flamingo,li2023otter,achiam2023gpt} have leveraged the formidable reasoning power of LLMs like LLaMA~\cite{zhang2023llama} and ChatGPT \cite{wu2023brief}. 
Inspired by the notable success of instruction tuning in LLMs~\cite{ouyang2022training, wang2022benchmarking, wang2022self}, the field of MLLMs is increasingly focusing on incorporating instruction-following data to further enhance models' understanding in downstream tasks. Pioneering work in this field encompasses LLaVA~\cite{liu2023visual} and InstructBLIP~\cite{dai2023instructblip}.
Recently, LLaVA-1.5~\cite{liu2023improved} has built upon LLaVA's framework, refining its instructions to improve performance across a wider range of VQA tasks. 
Recently, \citet{he2023continual} explored instruction tuning in the continual learning of MLLMs. Our research diverges from this study in terms of task setting, data assessment, and methodology. For more detailed discussions, please refer to \autoref{app:related}. \looseness=-1

\textbf{Catastrophic Forgetting.} 
(1) \textbf{Catastrophic Forgetting in LLMs:} Recently, there has been a growing focus on addressing the problem of overfitting in LLMs when fine-tuning on small datasets \cite{dong2021should, korbak2022controlling, howard2018universal, lee2019mixout, zhang2020revisiting} or after in-context learning \cite{alayrac2022flamingo, wang2023voyager}. The most advanced methods have endeavored to combat this challenge by employing techniques such as learning a sparse mask through loss optimization or randomly selecting masks~\cite{panigrahi2023task} and rescaling parameters to achieve model fusion~\cite{yu2023language}. Nevertheless, these methodologies have shown limited effectiveness when applied to MLLMs, owing to the intricate architecture and non-uniform weight distributions arising from their cross-modal nature \cite{sung2023ecoflap}.
(2) \textbf{Catastrophic Forgetting in MLLMs:} Few studies \cite{he2023continual,lin2023speciality,zhai2023investigating} explore this phenomenon in MLLMs. \citet{he2023continual} delves into traditional continual learning, which is distinct from our focus. \citet{lin2023speciality} examines CLIP's performance in image classification tasks, not extending to text generation models like LLaVA. Meanwhile, \citet{zhai2023investigating} investigates the performance of LLaVA and InstructBLIP in image classification tasks, without exploring deeper into multimodal tasks. Crucially, \citet{lin2023speciality} and \citet{zhai2023investigating} mainly apply existing anti-forgetting methods to MLLMs, concentrating on analytical approaches without proposing novel solutions. 
In contrast, our work focuses on the fundamental problem of catastrophic forgetting in multimodal generation and reasoning tasks in MLLMs, and we are the first parameter-efficient work to tackle this challenge. For an overview of catastrophic forgetting in smaller models, please refer to \autoref{app:related}.  \looseness=-1


\section{Problem Formulation}

Given a Multi-modal Large Language Model $\mathcal{M}$ either pre-trained or demonstrating strong generalization capabilities on a series of tasks $\mathcal{P} = \{D_1, D_2, \cdots, D_n\}$, where the model parameters are denoted by $\boldsymbol{\Theta}_{\mathrm{pre}}$. Catastrophic forgetting in MLLMs is characterized by a significant degradation in performance on previously learned tasks after undergoing supervised fine-tuning (SFT) for a new and unseen task $\mathcal{T} = D^{\text{new}}$, which updates the model's parameters to $\boldsymbol{\Theta}_{\mathrm{sft}}$. To counteract this, we aim to design a fusion function $\mathcal{F}(\cdot)$, formulated as follows:
\begin{equation}
 \boldsymbol{\Theta}_{\mathrm{fusion}} = \mathcal{F}(\boldsymbol{\Theta}_{\mathrm{sft}}, \boldsymbol{\Theta}_{\mathrm{pre}}),   
 \label{eq:fusion}
\end{equation}
where $\boldsymbol{\Theta}_{\mathrm{fusion}}$ denotes the optimally fused parameters. The primary goal of $\mathcal{F}(\cdot)$ is to enhance the performance of $\mathcal{M}$ on a downstream task $\mathcal{T}$, as well as preserve the pre-trained knowledge on original tasks $\mathcal{P}$.

Formally, the optimization objective is to minimize the loss on $\mathcal{T}$, while ensuring minimal deviation from the original task performances, which can be expressed as:
\begin{equation}
\begin{aligned}
& \mathcal{L}_{\mathcal{T}}(\boldsymbol{\Theta}_{\text{fusion}}) -  \mathcal{L}_{\mathcal{T}}(\boldsymbol{\Theta}_{\text{sft}}) \leq \epsilon_\text{t}, \\
& \mathcal{L}_{\mathcal{P}}\left(\boldsymbol{\Theta}_{\mathrm{fusion}}\right) - \mathcal{L}_{\mathcal{P}}\left(\boldsymbol{\Theta}_{\mathrm{pre}}\right) \leq \epsilon_{p} ,
\end{aligned}
\label{eq:global_obj}
\end{equation}
where $\mathcal{L}_{\mathcal{T}}$ and $\mathcal{L}_{\mathcal{P}}$ denote the loss functions corresponding to the downstream task $\mathcal{T}$ and the pre-training tasks $\mathcal{P}$. 
 $\epsilon_t$ and $\epsilon_p$ denote the acceptable performance degradation margins for $\mathcal{T}$ and $\mathcal{P}$ post-fusion.

\begin{figure*}[t]
  \centering
    \includegraphics[width=6.4in, height=1.6in]{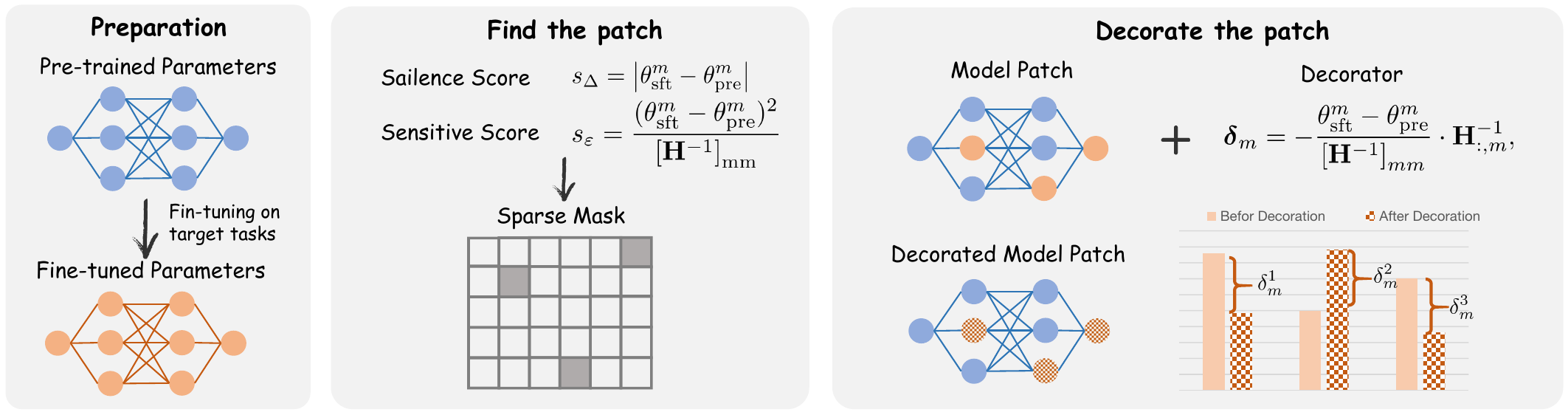}
  \caption{\textbf{Overall Framework of Model Tailor.}  Model Tailor consists of two primary steps. The first step in this process focuses on the identification of a ``model patch'', which is defined as a critical subset of fine-tuned parameters deemed essential for improving the model's effectiveness on a given target task. The second step is dedicated to applying compensatory adjustments, a methodological intervention designed to counterbalance any potential performance deficits that may arise from the exclusion of certain parameters during the fine-tuning phase.
  }
  \label{fig:framework}
  \vspace{-3mm}
\end{figure*}

\section{Method}
\label{sec:method}
\subsection{Overview of Model Tailor}
\label{sec:method_overview}
Inspired by the Lottery Ticket Hypothesis \cite{frankle2019lottery,li2023otter} and the Optimal Brain Surgeon~\cite{lecun1989optimal, hassibi1993optimal} (refer to \autoref{app:method} for related preliminaries), 
Model Tailor akin to a skilled tailor carefully selects and decorates patches of new fabric (fine-tuned parameters) and seamlessly integrating them onto the existing garment (pre-trained model). Model Tailor consists of two primary steps as illustrated in \autoref{fig:framework}: \looseness=-1

\textbf{Step 1: Identify model patch.} 
The initial step is akin to selecting the ideal piece of fabric, identifying a crucial subset of the model's fine-tuned parameters critical for enhancing target task performance. This crucial subset is termed as the \textit{``model patch"}.\looseness=-1

\textbf{Step 2: Decorate the patch.} After identifying the patch, we apply targeted compensation to this selected subset. 
The second step is designed to alleviate the loss rise on the target task due to the removal of other non-selected fine-tuned parameters. This compensatory step known as \textit{``patch decorator"}, is akin to adding embellishments to a garment, enhancing the model's capabilities on the target task.

Through the aforementioned two steps, Model Tailor can adeptly fine-tune the model's performance on the target task while meticulously preserving the core integrity of its pre-trained knowledge base. Formally, the fusion process in Model Tailor can be expressed as:
\begin{equation}
\begin{aligned}
   {\boldsymbol{\Theta}}_\text{fusion} &= \mathcal{F}(  {\boldsymbol{\Theta}}_\text{sft},  {\boldsymbol{\Theta}}_\text{pre})  \\
&= \boldsymbol{M} \odot ( {\boldsymbol{\Theta}}_\text{sft} + \boldsymbol{C} ) + (\boldsymbol{I} - \boldsymbol{M}) \odot {\boldsymbol{\Theta}}_\text{pre},
\end{aligned}
\label{eq:fusion_tailor}
\end{equation}
where $\boldsymbol{I}$ represents the identity matrix and $\boldsymbol{M}$ is a sparse binary mask used to identify the model patch, while $\boldsymbol{C}$ represents the patch decorator, referring to the compensation values for the parameters selected by $\boldsymbol{M}$.

\subsection{Layer-wise Objective of Model Tailor} 
Due to the large scale of $\mathcal{M}$, calculating masks and compensation values poses a considerable computational challenge. To mitigate this, we leverage a pruning-inspired strategy~\cite{frantar2022optimal, frantar2023sparsegpt}, breaking down Model Tailor into more manageable layer-wise sub-tasks. 
For each layer $\ell$ in $\mathcal{M}$, we define the layer function as $f_{\ell}\left(X_{\mathcal{T}}^{\ell}, {\boldsymbol{\Theta}}^{\ell}_\text{sft} \right)$, 
where $X_{\mathcal{T}}^{\ell}$ signifies the layer inputs for the target task and ${\boldsymbol{\Theta}}^{\ell}_\text{sft}$ represents the corresponding layer weights in layer $\ell$.
The expectation over the layer inputs $X_{\mathcal{T}}^{\ell}$ is usually approximated by taking the mean over a small set of $N$ input samples~\cite{frantar2022optimal}. 
Our goal is to refine ${\boldsymbol{\Theta}}^{\ell}_\text{fusion}$ to achieve performance as close as possible to the fully fine-tuned weights for the target task and fully pre-trained weights for the original tasks. 
Formally, we seek to optimize the weights ${\boldsymbol{\Theta}}^{\ell}_\text{fusion}$ such that they minimize any discrepancies in expected layer output, which can be expressed as:
\begin{equation}
\begin{aligned}
      \underset{{{\boldsymbol{\Theta}}^{\ell}_\text{fusion}}}{\arg\min}& \, \mathbb{E}_{\boldsymbol{X}_{\mathcal{T}}^{\ell}} \mathcal{L}\left(f_{\ell}\left(\boldsymbol{X}_{\mathcal{T}}^{\ell}, {\boldsymbol{\Theta}}^{\ell}_\text{sft} \right), f_{\ell}\left(\boldsymbol{X}_{\mathcal{T}}^{\ell}, {\boldsymbol{\Theta}}^{\ell}_\text{fusion} \right)\right),  \\
      \text {s.t.}& \, \mathbb{E}_{\boldsymbol{X}_{\mathcal{P}}^{\ell}} \mathcal{L}\left(f_{\ell}\left(\boldsymbol{X}_{\mathcal{P}}^{\ell}, {\boldsymbol{\Theta}}^{\ell}_\text{pre}\right), f_{\ell}\left(\boldsymbol{X}_{\mathcal{P}}^{\ell}, {\boldsymbol{\Theta}}^{\ell}_\text{fusion}\right)\right) < \epsilon ,
\end{aligned}
\label{eq:layer_obj}
\end{equation}
where ${\boldsymbol{\Theta}}^{\ell}_\text{fusion} = \mathcal{F}({\boldsymbol{\Theta}}^{\ell}_\text{sft},{\boldsymbol{\Theta}}^{\ell}_\text{pre})$. $\boldsymbol{X}_\mathcal{P}^{\ell}$ is the layer input on pre-trained tasks. $\epsilon$ is a threshold.  $\mathcal{L}(\cdot)$ represents a specific loss function, commonly employed as the squared loss in many works~\cite{hubara2021accurate, nagel2020up, wang2020towards}. The constraint imposed on this optimization ensures that the performance on the pre-training tasks $\mathcal{P}$ does not degrade beyond $\epsilon$. This optimization target aligns with the global objective introduced in \autoref{eq:global_obj}, adapting it to a layer-wise perspective with a focus on output consistency.\looseness=-1

It is worth noting that the practical implementation of \autoref{eq:layer_obj} faces obstacles due to the unavailability of pre-training data $\boldsymbol{X}_\mathcal{P}^\ell$, often restricted by proprietary or privacy considerations.
Therefore, we approximate this constraint by limiting the distance between fusion parameters ${\boldsymbol{\Theta}}^{\ell}_\text{fusion}$ and pre-trained parameters ${\boldsymbol{\Theta}}^{\ell}_\text{pre}$. In other words, our goal is to retain as many pre-trained parameters as possible to maintain the model's performance on the original task. Hence, the layer-wise optimization objective can be explicitly formulated as follows:
\begin{equation}
\begin{aligned}
 \underset{{{\boldsymbol{\Theta}}^{\ell}_\text{fusion}}}{\arg\min}& \left\|{\boldsymbol{\Theta}}^{\ell}_\text{sft} \mathbf{X}_{\mathcal{T}}^{\ell}-{\boldsymbol{\Theta}}^{\ell}_\text{fusion} \mathbf{X}_{\mathcal{T}}^{\ell}\right\|_2^2, \\
 \text{s.t.}& \, \|{\boldsymbol{\Theta}}^{\ell}_\text{fusion} - {\boldsymbol{\Theta}}^{\ell}_\text{pre}\|_1 < \eta ,
\end{aligned}
\label{eq:layer_obj_true}
\end{equation}
where $\eta$ represents a constraint parameter that denotes the allowable distance or deviation between the fusion parameters ${\boldsymbol{\Theta}}^{\ell}_\text{fusion}$ and the pre-trained parameters ${\boldsymbol{\Theta}}^{\ell}_\text{pre}$. This constraint ensures that the fusion parameters stay within a certain range of the pre-trained parameters.



\subsection{Identify Model Patch}

Building upon the introduction in \autoref{sec:method_overview}, we provide the formal definitions of model patch and layer patch:

\begin{definition}[Model Patch]
\label{def:model_patch}
Given a MLLM $\mathcal{M}$ with its pre-trained parameters ${\boldsymbol{\Theta}}_{\text{pre}}$ and fine-tuned parameters ${\boldsymbol{\Theta}}_{\text{sft}}$, A \textit{model patch} ${\boldsymbol{\Theta}}_{\text{patch}} \subseteq {\boldsymbol{\Theta}}_{\text{sft}}$, is identified through a binary mask $\boldsymbol{M} \in \{0, 1\}^{|\boldsymbol{\Theta}_\text{sft}|}$, defined as:
\begin{equation}
{\boldsymbol{\Theta}}_{\text{patch}} = \{ \theta_i | M_i = 1, \theta_i \in {\boldsymbol{\Theta}}_{\text{sft}} \},
\end{equation}
where $M_i$ denotes the $i$-th element of $\boldsymbol{M}$. ${\boldsymbol{\Theta}}_{\text{patch}}$ captures the most crucial subset of fine-tuned parameters for the target task, thereby facilitating the achievement of the global objective expressed in \autoref{eq:global_obj}.
\end{definition}

Then, we extend this concept to individual layers within the model, introducing the notion of a Layer Patch.

\begin{definition}[Layer Patch]
\label{def:layer_patch}
For each layer $\ell$ in an MLLM, characterized by layer-specific pre-trained parameters ${\boldsymbol{\Theta}}^{\ell}_{\text{pre}}$ and fine-tuned parameters ${\boldsymbol{\Theta}}^{\ell}_{\text{sft}}$, a layer patch ${\boldsymbol{\Theta}}^{\ell}_{\text{patch}} \subseteq {\boldsymbol{\Theta}}^{\ell}_{\text{sft}}$ is derived by $\boldsymbol{M}^{\ell} \in \{0, 1\}^{|{\boldsymbol{\Theta}}^{\ell}_{\text{sft}}|}$, expressed as:
\begin{equation}
{\boldsymbol{\Theta}}^\ell_{\text{patch}} = \{ \theta_\text{sft}^i | M^\ell_i = 1, \theta^i_\text{sft} \in {\boldsymbol{\Theta}}^\ell_{\text{sft}} \},
\end{equation}
with $M^\ell_i$ being the $i$-th element of $\boldsymbol{M}^\ell$. Similarly, ${\boldsymbol{\Theta}}^{\ell}_{\text{patch}}$ enables the achievement of the layer-wise optimization objective expressed in \autoref{eq:layer_obj_true}.
\end{definition}


We are naturally led to a crucial question: how do we identify this layer patch, or more specifically, how do we determine the mask $\boldsymbol{M}^\ell$? This inquiry propels us to investigate two principal mask selection strategies.




\textbf{Sensitivity-based Score.}  Drawing from the principles of OBS, we introduce a sensitivity based score, $s_\varepsilon$, as a measure to evaluate the impact of reverting a fine-tuned parameter $\theta_{\text{sft}}$ to its pre-tuning state $\theta_{\text{pre}}$ on the target task performance. We present~\autoref{theorem:sensitive} proved in \autoref{app:proof}, which offers second-order analytical lens to evaluate the significance of each parameter.

\begin{theorem} 
\label{theorem:sensitive} 
Consider a layer $\ell$ within an MLLM $\mathcal{M}$, and let $\theta_m$ represent a parameter at index $m$. Altering $\theta_m$ from its fine-tuned state $\theta^m_{\text{sft}}$ to its pre-trained state $\theta^m_{\text{pre}}$, induces a increase $\Delta \mathcal{L}_\mathcal{T}$ in the model's loss for $\mathcal{T}$, quantified as:
\begin{equation}
\Delta \mathcal{L}_\mathcal{T} =\frac{(\theta^m_{\text{sft}} - \theta^m_{\text{pre}})^2}{\left[\mathbf{H}^{-1}\right]_\text{mm}},
\end{equation}
where $\mathbf{H}^{-1}$ denotes the inverse of the Hessian matrix and $\left[\mathbf{H}^{-1}\right]_{mm}$ is its $m$-th diagonal element. The Hessian matrix $\mathbf{H}=\nabla_{\theta^m_\text{sft}}^2 \mathcal{L}_{\mathcal{T}}$ takes into account the local geometry of the loss at a given point $\theta^m_\text{sft}$ assuming $\mathcal{L}_\mathcal{T}$ is twice differentiable.
\end{theorem}

The optimal selection of the parameter index $m$ that results in the smallest increase in loss. This increase in loss is quantified as the sensitivity-based statistic $s_\varepsilon$:
\begin{equation}
    s_\varepsilon = \frac{( \theta^m_\text{sft}-{\theta}_\text{pre}^m)^2}{2\left[\mathbf{H}^{-1}\right]_{mm}}.
    \label{eq:s_var}
\end{equation}
The value of $s_\varepsilon$ is indicative of how the adjustment of the $m$-th parameter influences the model's task-specific performance, with higher scores highlighting parameters of paramount importance to the model's efficacy.





\textbf{Salience-based Score.} Our second strategy employs a salience based score that measures the absolute deviation of fine-tuned parameters from their pre-trained counterparts, focusing on those that have been significantly altered to adapt to the specifics of the target task. This deviation is quantitatively captured for each parameter $\theta^m_\text{sft} \in \boldsymbol{\Theta}^\ell_\text{sft}$ within layer $\ell$, by computing:
\begin{equation}
s_\Delta = \left| \theta^m_{\text{sft}} - \theta^m_{\text{pre}} \right|,
\end{equation}
where $s_\Delta$ serves as an indicator of a parameter's adaptability. This metric underscores the parameters that have shown substantial responsiveness to the task adaptation process, thereby shedding light on their pivotal role in the model's learning evolution for the new task.

\textbf{Score Fusion.}  To discern the most impactful parameters for a layer $\ell$, we devise a hybrid importance score, $s_\text{h}$, synthesized from two above perspectives:

\begin{equation}
s = \omega \cdot \tilde{s}_\Delta + (1 - \omega) \cdot \tilde{s}_\varepsilon,
\label{eq:mask}
\end{equation}

where $\tilde{s}_\Delta$ and $\tilde{s}_\varepsilon$ represent the min-max normalized salience and sensitivity scores, respectively. $\omega$ is a weighting factor that adjusts the balance between the $\tilde{s}_\Delta$ and $\tilde{s}_\varepsilon$.

Let $\rho$ represent the predefined sparsity level, indicating the percentage of parameters to be removed. The process of determining the final mask $M^{\ell}$ for layer $\ell$ can be expressed as: $M^{\ell}_i = \mathbbm{1}\left\{s^i \geq s^\prime \right\}$. $\mathbbm{1}\{\cdot\}$ is the indicator function, assigning a value of 1 when the condition is met and 0 otherwise. $s^\prime$ denotes the score threshold, determined by the $\rho$-th highest score. By prioritizing parameters above the hybrid score threshold, we approximate the layer patch outlined in \textcolor{someorange}{Definition} \ref{def:layer_patch}, focusing solely on parameters vital for the target task's performance. 







\subsection{Decorate the Patch}

Once identifying the layer patch $\Theta^\ell_{\text{patch}}$, the subsequent step involves optimizing these selected parameters to ensure the model's performance on the target task remains robust. Then, we have the definition of the patch decorator, a crucial mechanism for enhancing the layer's effectiveness.

\begin{definition}[Patch Decorator]
\label{def:decorator}
Given a layer $\ell$ in an MLLM and the corresponding layer patch $\Theta^\ell_\text{patch}$ identified by the binary mask $M^\ell$,
the patch decorator $\delta^i$ is the adjustment for each selected parameter $\theta^i_{\text{sft}} \in \Theta^\ell_\text{patch}$, yielding the ajusted layer patch:

\begin{equation}
\widehat{\boldsymbol{\Theta}}^\ell_{\text{patch}} = \{ \theta_\text{sft}^i + \delta^i | \boldsymbol{M}^\ell_i = 1, \theta^i_\text{sft} \in {\boldsymbol{\Theta}}^\ell_{\text{sft}} \}.
\end{equation}
\end{definition}

The patch decorator is designed to compensate the increased loss in target performance resulting from excluding parameters not covered by the binary mask $\boldsymbol{M}^\ell$, i.e., $\Theta^\ell_{\text{sft}} \setminus \Theta^\ell_{\text{patch}}$.

\begin{table*}[t]
  \caption{\textbf{H-score and average performance on InstructBLIP (Vicuna-7B) with a sparsity ratio $\rho=10\%$.} ``\#Params" refers to the number of parameters modified. The optimal and sub-optimal results are denoted by boldface and underlining. \looseness=-1}
  \vspace{-2.5mm}
  \centering
  \resizebox{1\linewidth}{!}{
  \begin{tabular}{l|l|cccccccc|c|>{\columncolor{tabhighlight}}c >{\columncolor{tabhighlight}}c}

      \bottomrule

      \toprule
      \multirow{2}{*}{Method} &  \multirow{2}{*}{\#Params} &  \multicolumn{8}{c}{\textbf{Pre-trained tasks}} & \multicolumn{1}{c}{\textbf{Target task}}\\
      & &
      \multicolumn{1}{c}{\textcolor{someblue}{COCO}} &
      \multicolumn{1}{c}{\textcolor{someblue}{NoCaps-in}} &
      \multicolumn{1}{c}{\textcolor{someblue}{NoCaps-near}} &
      \multicolumn{1}{c}{\textcolor{someblue}{NoCaps-out}} &
      \multicolumn{1}{c}{\textcolor{someblue}{OKVQA}} &
      \multicolumn{1}{c}{\textcolor{someblue}{AOKVQA}} &
      \multicolumn{1}{c}{\textcolor{someblue}{VQAv2}} &
       \multicolumn{1}{c}{\textcolor{someblue}{GQA}} &
      \multicolumn{1}{c}{\textcolor[rgb]{0.773,0.353,0.067}{Flickr30k}} &
      \multicolumn{1}{c}{Avg} &
      \multicolumn{1}{c}{Hscore} 
      \\  
      
      \midrule
      {Zero-shot} & - & 143.1 & 117.7 & 123.5&121.9& 58.45 &  58.24 & 76.66 & 49.18  & 82.8 & 92.39 & 87.87    \\
      \midrule
  {fine-tune}& 288M & 114.2 & 101.9& 113.2& 112.3& 54.47 & 55.30& 74.26& 47.29 &  101.8 &  86.08&  92.12       \\
  {DARE} & 28.8M & 114.1 & 103.3& 113.5&109.2 & {58.07} & 55.89 &  {76.65} & {49.19} &  {101.3} & 86.8 & 92.43             \\
  {Grafting} & 28.8M &  {127.9} & {112.6}& {121.5} & 116.3& 53.58& 54.58 &71.90 & 47.01  & {97.9} & \underline{89.25} &  \underline{92.78}                 \\
  {Ours} &28.8M &  {139.8} & {115.8} &  {124.3} & {123.4} & {57.63}  & {57.84} & {76.19} & {48.58}  & 94.4 & \textbf{93.10} & \textbf{93.67}      \\
  \bottomrule

  \toprule

   \multirow{2}{*}{Method} &  \multirow{2}{*}{\#Params} &  \multicolumn{8}{c}{\textbf{Pre-trained tasks}} & \multicolumn{1}{c}{\textbf{Target task}}\\
      & &
      \multicolumn{1}{c}{\textcolor{someblue}{COCO}} &
      \multicolumn{1}{c}{\textcolor{someblue}{NoCaps-in}} &
      \multicolumn{1}{c}{\textcolor{someblue}{NoCaps-near}} &
      \multicolumn{1}{c}{\textcolor{someblue}{NoCaps-out}} &

      \multicolumn{1}{c}{\textcolor{someblue}{OKVQA}} &
      \multicolumn{1}{c}{\textcolor{someblue}{AOKVQA}} &
      \multicolumn{1}{c}{\textcolor{someblue}{VQAv2}} &
       \multicolumn{1}{c}{\textcolor{someblue}{Flickr30k}} &
      \multicolumn{1}{c}{\textcolor[rgb]{0.773,0.353,0.067}{GQA}} &
      \multicolumn{1}{c}{Avg} &
      \multicolumn{1}{c}{Hscore} 
      \\  
      
      \midrule
      {Zero-shot} & - & 143.1 & 117.7 & 123.5&121.9&  58.45 & 58.24 &  76.66 & 82.8 & 49.18 & 92.39& 65.44     \\
      \midrule
  {fine-tune}  & 288M & 109.8 & 94.9&99.0 &100.4 & 38.32 & 41.13 &67.85 & 66.6 & 59.81 & 75.31 & 67.42        \\
  {DARE} &  28.8M & {140.2} &  {115.8}& {119.3} &118.5  & {58.01} & {57.25} & {74.74}& {81.8}&  {50.91} &  \underline{90.72} & \underline{66.46}     \\
  {Grafting} & 28.8M  &138.5 &114.3 &119.2 & {119.5} &55.09 & {57.06} &74.36&80.3 & 50.82& 89.90 & 66.16       \\
  {Ours} & 28.8M &  {141.8} &  {116.5} & {121.8} & {120.7} & {58.33} & 53.98 & {76.28} &  {82.7}  & {56.15} & \textbf{92.02} & \textbf{71.00}   \\
  \bottomrule

  \toprule

      


  \end{tabular}
  }
  \label{tab:main_instructblip}
  \vspace{-3mm}
  \end{table*}

 \autoref{theorem:sensitive} delineates the change in loss resulting from the alteration of a parameter from its fine-tuned state to its pre-trained state. Building upon this, we present \autoref{theorem:compensate} proved in \autoref{app:proof}, which specifies the compensatory values for the remaining parameters.

\begin{theorem}
\label{theorem:compensate}
Given the alteration of a parameter from its fine-tuned state, $\theta_m^{\text{ft}}$, to its pre-trained state, $\theta_m^{\text{pre}}$, the optimal perturbation $\Delta \Theta$ for each retained fine-tuned parameters is calculated as follows:
\begin{equation}
\Delta \boldsymbol{\Theta}_m^* = -\frac{\theta^m_{\text{sft}} - \theta^m_{\text{pre}}}{\left[\mathbf{H}^{-1}\right]_{mm}} \cdot \mathbf{H}_{:,m}^{-1},
\end{equation}
where $\mathbf{H}^{-1}$ denotes the inverse of the Hessian matrix and $\mathbf{H}_{:,m}^{-1}$ signifies the  $m$-th column of $\mathbf{H}^{-1}$. 
\end{theorem}


\autoref{theorem:compensate} serves as the foundational pillar for determining the patch decorator in Model Tailor. In fact, the patch decorator $\delta^i$ represents the $i$-th value within the vector $\Delta \boldsymbol{\Theta}_m^*$. By applying $\Delta \boldsymbol{\Theta}_m^*$ to the selected sparse parameters, our method ensures a minimally disruptive integration of the fine-tuned parameters with the pre-trained model. 

Additionally, Model Tailor can be extended to multi-task scenarios to absorb and integrate knowledge from multiple tasks, details described in \autoref{app:method}.

\begin{figure}[h]
  \centering
    \includegraphics[width=3.4in]{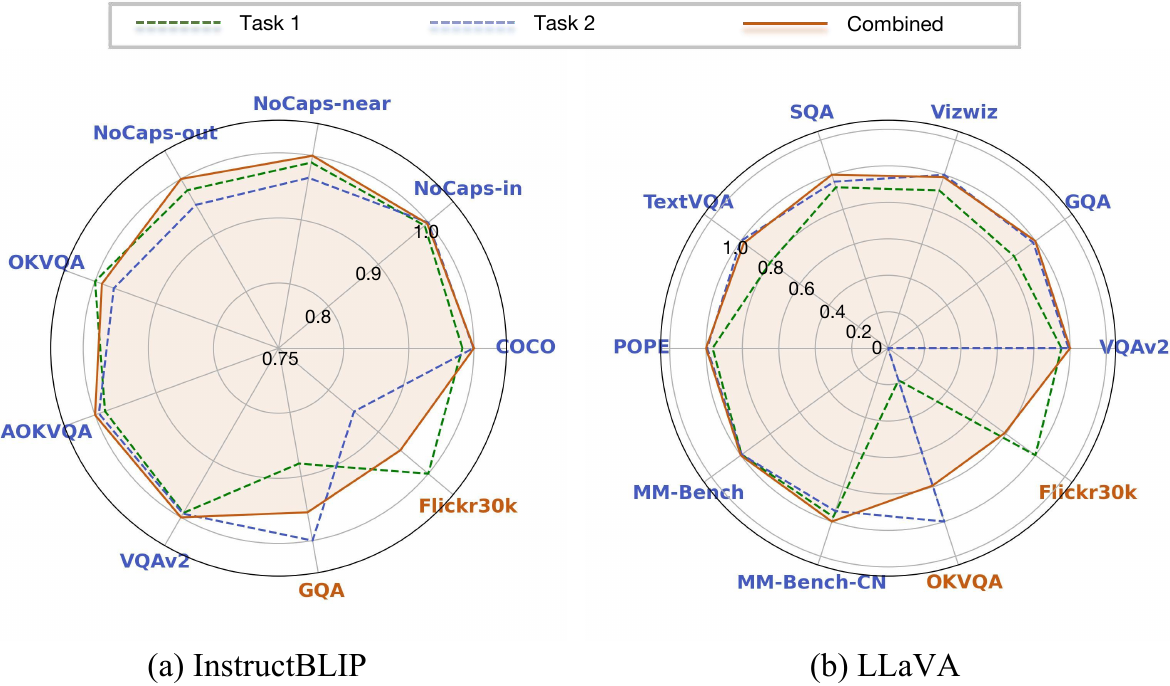}
  \caption{\textbf{Model Tailor on Multi-Task Scenario. } ``\textit{Performance oscillations}" where models exhibit dips in efficacy on one task after fine-tuning on another, which are effectively bridged by Model Tailor's multi-task fusion.
  }
  \label{fig:multitask}
  \vspace{-5mm}
\end{figure}

\section{Experiments}
\label{sec:experiments}
\subsection{Experimental Setup}

\textbf{Architectures and Datasets.}
Given the representativeness of InstructBLIP (Vicuna-7B) and LLaVA-1.5 (Vicuna-7B) in terms of structure and the scope of fine-tuned parameters—where the former fine-tunes the projector and the latter adjusts both the projector and LLM, we evaluate Model Tailor on these two models.
For InstructBLIP, following \cite{dai2023instructblip}, we engage with datasets including COCO Caption~\cite{lin2014microsoft}, NoCaps (in, near, out)~\cite{agrawal2019nocaps}, OKVQA~\cite{marino2019ok}, AOKVQA~\cite{schwenk2022okvqa}, GQA~\cite{hudson2019gqa}, VQAv2~\cite{goyal2017making}, and Flickr30k~\cite{young2014image}. Specifically, we fine-tune the models on Flickr30k and GQA - datasets associated with image captioning and VQA tasks and not encountered during the pre-training phase of InstructBLIP. 
Similarly, for LLaVA-1.5, in line with~\cite{liu2023improved}, the datasets involved are VQAv2, GQA, Vizwiz~\cite{gurari2018vizwiz}, SQA~\cite{lu2022learn}, TextVQA~\cite{singh2019towards}, POPE~\cite{li2023evaluating}, MM-Bench~\cite{liu2023mmbench}, and MM-Bench-CN~\cite{zhang2023internlm}. 
We fine-tune LLaVA on Flickr30k and OKVQA tasks, both distinct from its pre-training exposure, and assess its performance on the remaining datasets.

  \begin{table*}[t]
    \caption{\textbf{H-score and average performance on LLaVA-1.5 (Vicuna-7B) with a sparsity ratio $\rho=10\%$.} ``\#Params" refers to the number of parameters modified. The optimal and sub-optimal results are denoted by boldface and underlining. \looseness=-1}
    \vspace{-3mm}
  \label{tab:main_llava}
  \centering
  \resizebox{1\linewidth}{!}{
  \begin{tabular}{l|l|cccccccc|c|>{\columncolor{tabhighlight}}c >{\columncolor{tabhighlight}}c}

      \bottomrule
      
      \toprule
      \multirow{2}{*}{Method} &  \multirow{2}{*}{\#Params} &  \multicolumn{8}{c}{\textbf{Pre-trained tasks}} & \multicolumn{1}{c}{\textbf{Target task}}\\
      & &
      \multicolumn{1}{c}{\textcolor{someblue}{VQAv2}} &
      \multicolumn{1}{c}{\textcolor{someblue}{GQA}} &
      \multicolumn{1}{c}{\textcolor{someblue}{VizWiz}} &
      \multicolumn{1}{c}{\textcolor{someblue}{SQA}} &
      \multicolumn{1}{c}{\textcolor{someblue}{TextVQA}} &
      \multicolumn{1}{c}{\textcolor{someblue}{POPE}} &
      \multicolumn{1}{c}{\textcolor{someblue}{MM-Bench}} &
       \multicolumn{1}{c}{\textcolor{someblue}{MM-Bench-CN}} &
      \multicolumn{1}{c}{\textcolor[rgb]{0.773,0.353,0.067}{Flickr30k}} &
      \multicolumn{1}{c}{Avg} &
      \multicolumn{1}{c}{Hscore} 
      \\  
      
      \midrule
      {Zero-shot}  & -& 78.54 & 61.94& 50.0& 66.8& 58.27&85.9 & 64.3 & 58.3&  3.5 & 58.62& 6.64      \\
    \midrule
  {fine-tune} & 2.7B & 68.61 & 49.01 &27.24 & 63.86 &40.03 & 79.73& 59.02&50.17 &  77.1 & 56.42 &63.40      \\
  {DARE}  & 273M& {78.12} & {59.25}& {48.9}&64.92 & {57.17}& {84.86}&{64.77}& {57.47}& 25.6 & 60.12 & 36.64           \\
  {Grafting}  & 273M& {74.48} &{58.28} &{43.16} &{66.82} &{52.56} &80.35& {64.52} & 55.49 &{58.2}  & \underline{61.56} & \underline{60.03}              \\
  {Ours}  & 273M&73.21 &52.49 &42.28 & {67.15} &43.89 & {82.88}& 63.40 & {56.15} & {75.4}  & \textbf{61.87} &\textbf{66.94}         \\
  \bottomrule

  \toprule

   \multirow{2}{*}{Method} & \multirow{2}{*}{\#Params} &  \multicolumn{8}{c}{\textbf{Pre-trained tasks}} & \multicolumn{1}{c}{\textbf{Target task}}\\
      & &
       \multicolumn{1}{c}{\textcolor{someblue}{VQAv2}} &
      \multicolumn{1}{c}{\textcolor{someblue}{GQA}} &
      \multicolumn{1}{c}{\textcolor{someblue}{VizWiz}} &
      \multicolumn{1}{c}{\textcolor{someblue}{SQA}} &
      \multicolumn{1}{c}{\textcolor{someblue}{TextVQA}} &
      \multicolumn{1}{c}{\textcolor{someblue}{POPE}} &
      \multicolumn{1}{c}{\textcolor{someblue}{MM-Bench}} &
       \multicolumn{1}{c}{\textcolor{someblue}{MM-Bench-CN}} &
      \multicolumn{1}{c}{\textcolor[rgb]{0.773,0.353,0.067}{OKVQA}} &
      \multicolumn{1}{c}{Avg} &
      \multicolumn{1}{c}{Hscore} 
      \\  
      
      \midrule
      {Zero-shot}   & -& 78.54 & 61.94& 50.0& 66.8& 58.27&85.9 & 64.3 & 58.3& 0.14 &58.24 &0.28       \\
      \midrule
  {fine-tune} &2.7B & 69.1& 48.61&30.35 &41.03 &42.13 & 72.33 &32.79 &43.47 & 46.27& 47.34 &46.87     \\
  {DARE}  & 273M& {78.04} &  {61.65} & {49.19} & {67.58}& {57.91} & {86.44}  & {65.03}& {58.16}& 0.83 & 58.31 & 1.64           \\
  {Grafting}  & 273M&75.23 & 58.42& 43.27&67.26 &53.51 &85.29  & 62.16& 54.42& 30.8 &\underline{58.93}& \underline{41.25}              \\
  {Ours}  & 273M&  {76.25} & {60.39}&  {46.49}& {69.51} & {54.88} & {85.44} & {63.32}& 54.21&  {38.1} & \textbf{60.95}  &\textbf{47.71}   \\
  \bottomrule

  \toprule

  \end{tabular}
  }
  \vspace{-5mm}
  \end{table*}

  \begin{figure}[tpb]
  \centering
    \includegraphics[width=3.4in]{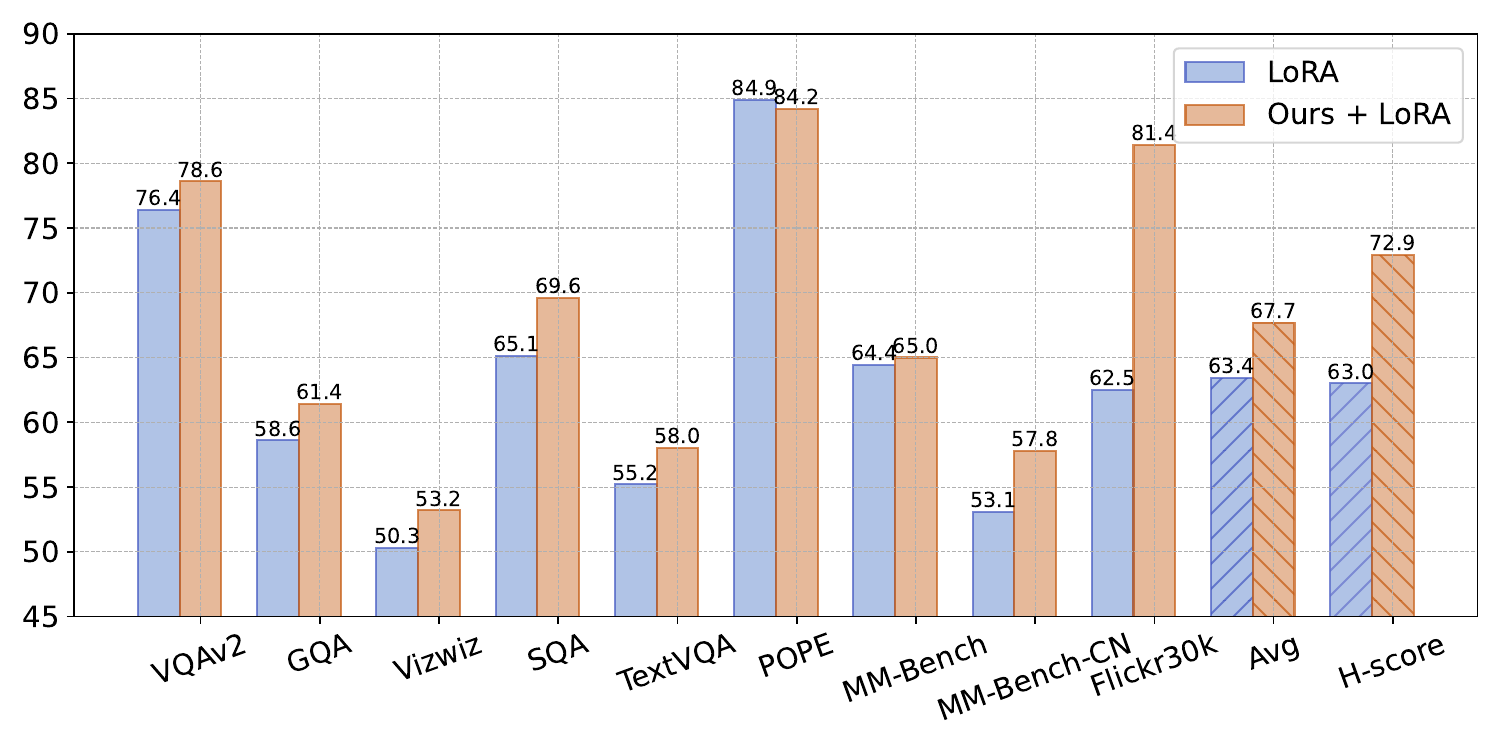}
   \vspace{-3mm}
  \caption{\textbf{Combination with LoRA on LLaVA.} The application of Model Tailor to LLaVA-1.5 fine-tuned using LoRA yields significant performance improvements across various datasets, indicating that Model Tailor’s post-training refinement complements LoRA.
  }
  \label{fig:lora_llava}
  \vspace{-5mm}
\end{figure}

\textbf{Compared Baselines.}
 We compare Model Tailor against three methods including:  (a) \textbf{Standard Fine-Tuning}: For InstructBLIP, the fine-tuning process targets the Q-Former, a connector bridging the vision encoder and the LLM, involving an extensive parameter set of approximately 288M. For LLaVA-1.5, fine-tuning is applied to both the MLP mapper and the final 12 layers of the LLM, constrained by GPU memory limits, involving a significant total of 2.7 billion parameters.
 (b) \textbf{Model Grafting}~\cite{panigrahi2023task}: This strategy optimizes the target task loss to derive a mask, which identifies the most pivotal parameters for the task at hand. 
 (c) \textbf{Drop \& Rescale (DARE)}~\cite{yu2023language}: In this approach, parameters from the fine-tuned model are randomly selected and rescaled to maintain the model's performance on both the target task and other tasks.

 \textbf{Evaluation Metrics.} We utilize the arithmetic and harmonic means of performance across pre-trained and target tasks. For more details, please refer to \autoref{app:experments}.

 \textbf{Implementation Details.} Given the computational intensity of calculating the inverse Hessian matrix, our method incorporates an optimization technique derived from SparseGPT~\cite{frantar2023sparsegpt}, adept at handling the complexities of large-scale models, thereby ensuring the practical feasibility of Model Tailor. Please see \autoref{app:experments} for detailed analysis and other implementation details.




\begin{table}[!t]
  \scriptsize
 
  \caption{ \textbf{Ablation Study of Model Tailor.} The effectiveness of decorator is critical on InstructBLIP and LLaVA.}
   \vspace{-2mm}
  \label{tab:delta_weight}
  \centering
  \resizebox{\linewidth}{!}
  {
  \renewcommand{\arraystretch}{0.9}
  \begin{tabular}{l|cc|>{\columncolor{tabhighlight}}c >{\columncolor{tabhighlight}}cc}
  \bottomrule

  \toprule
  & \multicolumn{4}{c}{InstructBLIP on Flickr30k} \\
  & 
  \multicolumn{1}{c}{\textcolor{someblue}{Pre-trained Tasks}} &
    \multicolumn{1}{c}{\textcolor{someorange}{Tagrte Task}} &
    \multicolumn{1}{c}{Avg} &
    \multicolumn{1}{c}{H-score} 
  \\
  \midrule
  Before Decoration & 90.75 & 93.21  & 91.02  & 91.96  \\
  After Decoration & 92.94 &  94.40 &  \textbf{93.10} & \textbf{93.67}  \\
  \bottomrule

    \toprule
  & \multicolumn{4}{c}{LLaVA on Flickr30k} \\
  & 
  \multicolumn{1}{c}{\textcolor{someblue}{Pre-trained Tasks}} &
    \multicolumn{1}{c}{\textcolor{someorange}{Target Task}} &
    \multicolumn{1}{c}{Avg} &
    \multicolumn{1}{c}{H-score} 
  \\
  \midrule
  Before Decoration &58.26  & 71.58  & 59.74  & 64.24  \\
  After Decoration &60.18  &  75.40 &  \textbf{61.87} & \textbf{66.94}  \\
  \bottomrule

\toprule
  \end{tabular}}
  \vspace{-5mm}
\end{table}



\begin{figure*}[t]
  \centering
    \includegraphics[width=6.8in]{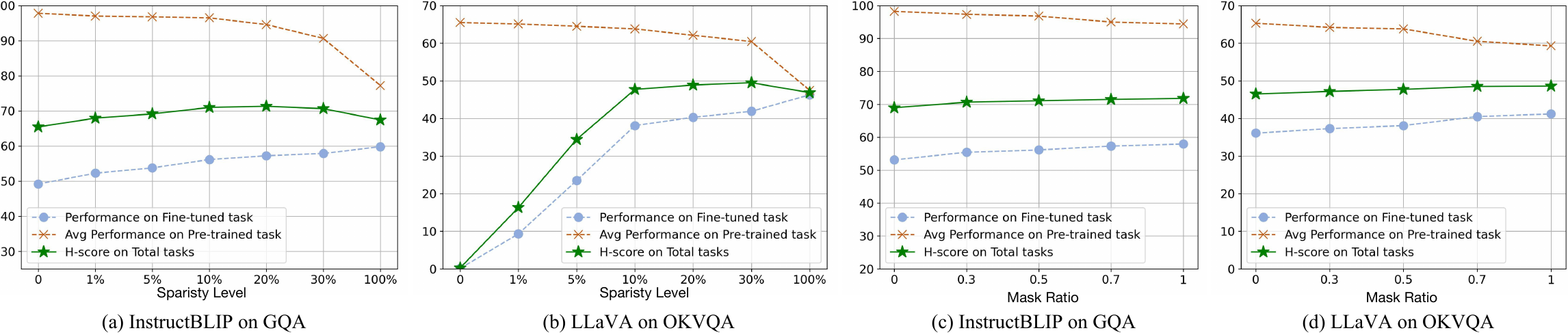}
  \caption{\textbf{Comparison the performance of Model Tailor in various settings.} (a-b) Results of InstructBLIP and LLaVA-1.5 at varying sparsity levels. (c-d) Results of InstructBLIP and LLaVA-1.5 with different mask proportions.
  }
  \label{fig:sparsity_mask}
  \vspace{-3mm}
\end{figure*}

\subsection{Main Results on Single-Task setting}


\autoref{tab:main_instructblip} and \autoref{tab:main_llava} illustrate that Model Tailor substantially mitigates catastrophic forgetting in MLLMs, outperforming current fine-tuning and forgetting mitigation methods in both InstructBLIP and LLaVA frameworks under a sparsity level of 10\%. We provide further important observations:

\vspace{-1mm}
\begin{itemize}[leftmargin=*]
    \item \textbf{Generalization capabilities of MLLMs raise significant concerns, particularly in adapting to unfamiliar tasks.} This is highlighted by the zero-shot performance of LLaVA on Flickr30k, and OKVQA is virtually null as shown in \autoref{tab:main_llava}. 
    Furthermore, subsequent fine-tuning to improve performance on such tasks often leads to a detrimental impact on the proficiency of pre-trained tasks. For example, LLaVA's performance on Vizwiz, as shown in \autoref{tab:main_llava}, falls sharply from a baseline score of 50.0 to 27.24 and 30.35 after fine-tuning for Flickr30k and OKVQA, respectively. Similarly, the performance of InstructBLIP on COCO, after undergoing fine-tuning, also experiences a significant drop, descending from a high of 143.1 to 114.2 and 109.8, as reported in \autoref{tab:main_instructblip}. 
    \item \textbf{Existing approaches to prevent forgetting demonstrate limitations when applied to MLLMs.} DARE, except in the case of InstructBLIP on Flickr30k, fails to sustain performance on target tasks, with LLaVA's performance on OKVQA dropping back to baseline after DARE's selection, as presented in \autoref{tab:main_llava}. Grafting, while offering some preservation of performance on target tasks, still leads to considerable forgetting on tasks for which the model was originally trained.
    \item \textbf{Model Tailor adeptly optimizes for specific tasks while preserving performance on pre-trained tasks.} It achieves superior H-scores and average metrics, highlighting its effectiveness in balancing focused enhancements with foundational robustness. This dual optimization is consistently reflected across InstructBLIP and LLaVA, confirming the method's comprehensive adaptability and excellence in the field.
\end{itemize}


\subsection{Main Results on Multi-Task Setting}

\autoref{fig:multitask} presents Model Tailor's adeptness in multi-task environments, utilizing relative performance indicators for clarity.The comparisons are conducted under a consistent sparsity level: each single task employs a sparsity of 10\% and the multi-task fusion combines from two 5\% sparsity tasks.
The radial plots disclose an apparent oscillation in performance—models fine-tuned for one task show dips in efficacy when assessed on another, creating what can be termed as ``\textit{performance oscillations}". However, when Model Tailor's multi-task fusion is applied, these \textit{performance oscillations} are seamlessly bridged. This not only rectifies the disparities but also leads to superior performance on the pre-trained tasks, suggesting that our method's integration of diverse knowledge domains significantly bolsters the model's robustness. Detailed quantitative assessments are provided in \autoref{tab:multitask_instructblip} and \autoref{tab:multitask_llava} in \autoref{app:experments}, which further substantiate the visual observations from \autoref{fig:multitask}.

\subsection{Synergy with Parameter-Efficient Method}

Model Tailor enhances the efficacy of existing parameter-efficient fine-tuning strategies such as LoRA~\cite{hu2021lora}. LoRA freezes pre-trained model weights and introduces trainable rank decomposition matrices, reducing parameters for downstream tasks. 
Specifically, we select and modify parameters on the model fine-tuned with LoRA. This synergy, shown in \autoref{fig:lora_llava}, particularly benefits the LLaVA fine-tuned on Flickr30k, where applying Model Tailor results in notable gains across a suite of datasets. Specifically, on Flickr30k, we achieved an 18.9-point increase, and the overall H-score improved by 9.9\%.  These findings imply that a subset of parameters modified during LoRA's fine-tuning could be extraneous, which Model Tailor adeptly identifies and excludes. Moreover, the Model Tailor's post-training refinement complements the fine-tuning conducted during LoRA, leading to improved model synergy. For a detailed comparison, \autoref{tab:lora_llava} in \autoref{app:experments} provides a performance breakdown, highlighting the benefits of our approach in two fine-tuning scenarios for LLaVA. \looseness=-1

\subsection{Ablation Study}

\textbf{Effectiveness of Decoration.}
\autoref{tab:delta_weight} elucidates the function of the decoration phase in Model Tailor, distinguishing between the pre-decorated state, ``Before Decoration", where only mask selection has occurred, and ``After Decoration", which involves weight compensation of the mask-selected parameters. The table clearly illustrates the efficacy of the decoration process, which significantly elevates model performance. This phase is crucial for ensuring that it adapts effectively to target tasks. More visualization results are provided in \autoref{fig:decorator} in \autoref{app:experments}.

\textbf{Influence of Sparsity Level.}
\autoref{fig:sparsity_mask} (a-b) examine the effects of sparsity on Model Tailor's performance. A pattern emerges where increasing sparsity correlates with enhanced performance on pre-trained tasks and a decline in target tasks. A performance plateau is observed across all tasks once sparsity exceeds 10\%. This plateauing effect is accentuated in LLaVA, particularly due to its subpar generalization on target tasks. The significant gains in performance from 0 to 10\% sparsity highlight the presence of superfluous parameters, suggesting that fine-tuning incorporates a considerable amount of redundancy.

\textbf{Trade-off on Mask Selection.}
The impact of the two masking mechanisms used in model patch identification process is further investigated. As depicted in \autoref{fig:sparsity_mask} (c-d), we study the influence of changes in the $\omega$ parameter in \autoref{eq:mask}. An increased $\omega$ value, giving more weight to the salience-based score, correlates with improved performance on target tasks. Conversely, a reduced 
$\omega$ value, which gives more weight to the sensitivity-based score, enhances performance on pre-trained tasks. These findings suggest that salience-based scores are instrumental in honing the model's capabilities for new tasks, while sensitivity-based scores fortify the foundational knowledge of pre-trained tasks. A balanced combination of these scores yields a harmonious state where both new and pre-existing task performances are optimized. \looseness=-2

\section{Conclusion}
In conclusion, this study addresses the critical challenge of catastrophic forgetting in fine-tuning MLLMs. We have introduced Model Tailor, which strategically preserves the integrity of pre-trained parameters while selectively fine-tuning others for task-specific optimization. Employing a hybrid strategy, Model Tailor identifies a sparse mask that pinpoints the most impactful parameters for adaptation. This mask can be flexibly combined across multiple tasks. Subsequent weight compensation on this curated subset significantly enhances target performance. The empirical results further confirm the effectiveness of Model Tailor.

\section*{Impact Statements}
This paper presents work whose goal is to advance the field of Machine Learning. There are many potential societal consequences of our work, none of which we feel must be specifically highlighted here.

\nocite{langley00}

\bibliography{example_paper}
\bibliographystyle{icml2024}

\newpage
\appendix
\onecolumn

\icmltitle{Appendix of Model Tailor: Mitigating Catastrophic Forgetting in MLLMs}

In this appendix, we provide the details omitted in the main text, offering additional analyses, proofs, and discussions.
\begin{itemize}
\item \autoref{app:related}: Additional discussions on related works (cf. \autoref{sec:related} of the main paper).
\item \autoref{app:method}: Preliminaries and its extension to multi-task scenarios to our method (cf.\autoref{sec:method} of the main paper).
\item \autoref{app:proof}: Detailed proofs of \autoref{theorem:sensitive} and \autoref{theorem:compensate} (cf. \autoref{sec:method} of the main paper).
\item \autoref{app:experments}: Comprehensive experimental details and further results. (cf. \autoref{sec:experiments} the main paper).
\end{itemize}

\section{More Details about Related Work}
\label{app:related}

\textbf{Difference with Continue Instruction Tuning.} Recently, \cite{he2023continual} has applied instruction tuning to the continual learning of MLLMs. Our work differs from this study in three ways:
(1) While ~\cite{he2023continual} focus on sequentially tasks and measuring the forgetting of older tasks when learning new tasks, our focus is on the forgetting of the pre-trained MLLM after fine-tuning specific tasks.
(2) We do not require measuring task similarity or acquiring original task data, making our method more applicable to real-world scenarios.
(3) Our post-training method is more flexible and parameter-efficient, as we only select a small number of parameters, avoiding full-model fine-tuning.

\textbf{Catastrophic Forgetting in Small Models.} This phenomenon is predominantly observed in continuous learning, which revolve around the sequential handling of tasks while quantifying the extent of forgetting of older tasks when new tasks are introduced. ~\cite{tao2020few, masana2022class}. Current methods primarily focus on full-mdoel fine-tuning, such as the approach by \cite{xuhong2018explicit, ritter2018online, aljundi2018memory, schwarz2018progress}, which imposes a penalty on parameter changes for new tasks. Meanwhile, \cite{yang2023neural} draws inspiration from the neural collapse and introduce a distribution-aware loss. However, these strategies are not suitable for large-scale pre-trained models.

\section{More Details about Method}
\label{app:method}
\subsection{Preliminary}
\label{app:method_preliminary}
\textbf{Lottery Tickets Hypothesis.} 
The Lottery Ticket Hypothesis~\cite{frankle2019lottery} proposes an intriguing concept in neural networks, suggesting that within a dense, randomly-initialized network lies a smaller, potentially powerful subnetwork. This subnetwork, when trained in isolation, can achieve comparable or superior test accuracy to the original network, within a similar or reduced number of training iterations.

To elaborate, consider a dense feed-forward neural network denoted as $f(x; \theta)$, where $\theta$ represents the network's initial parameters, following a distribution $\theta_0 \sim \mathcal{D}_\theta$. Upon training this network using stochastic gradient descent (SGD) on a given dataset, it attains a minimum validation loss $l$ at a specific iteration $j$, alongside a test accuracy of $a$. The hypothesis further explores the scenario of training the network $f(x; m \odot \theta)$, where $m \in \{0,1\}^{|\theta|}$ acts as a binary mask applied to the network's parameters. This setup maintains the initial parameter state as $m \odot \theta_0$. Through SGD optimization on the identical dataset, while keeping $m$ constant, the network is expected to reach a minimum validation loss $l'$ by iteration $j'$, achieving a test accuracy $a'$.
There exists at least one mask $m$ that allows for training within the same or fewer iterations ($j' \leq j$), achieving similar or better accuracy ($a' \geq a$), while significantly reducing the number of parameters utilized ($\|m\|_0 \ll |\theta|$). 
This hypothesis highlights the existence of optimal subnetworks within larger networks that can perform equally well with a fraction of the parameters.

\textbf{Optimal Brain Surgeon.} The Optimal Brain Surgeon framework~\cite{lecun1989optimal,hassibi1993optimal} focuses on pruning weights from a densely connected neural network with the least increase in loss. It hinges on a meticulous calculation involving the Hessian matrix, $\mathbf{H}$, of the loss function. This matrix encapsulates the second-order partial derivatives of the loss w.r.t. the network's weights, offering deep insights into the loss landscape around the current model parameters.

The decision to prune a particular weight, $w_p$, is based on minimizing the ratio $\frac{w_p^2}{\left[\mathbf{H}^{-1}\right]_{pp}}$, where $\left[\mathbf{H}^{-1}\right]_{pp}$ refers to the $p$-th diagonal entry of the inverse Hessian matrix. This approach ensures the selected weight for pruning is the one whose removal incurs the minimal increase in the loss, effectively identifying the least critical connections in terms of the model's performance. Moreover, OBS proposes a method to adjust the remaining weights in the network to compensate for the pruned weight. The adjustment for the remaining weights, $\boldsymbol{\delta}{\boldsymbol{p}}$, is determined by $-\frac{w_p}{\left[\mathbf{H}^{-1}\right]_{pp}} \cdot \mathbf{H}_{:, p}^{-1}$, where $\mathbf{H}_{:, p}^{-1}$ is the $p$-th column of the inverse Hessian. This compensatory update aims to mitigate the loss increase caused by pruning, ensuring the network maintains its performance as much as possible post-pruning.



\subsection{Extension to Multi-Task Scenario}
\label{app:method_multitask}
Expanding upon the above discussions centered around fine-tuning for a singular target task, our methodology exhibits versatility through its adaptability to multi-task scenarios.
Formally, consider $m$ target tasks, denoted as $\mathcal{T}=\{D_1,D_2, ...,D_m\}$, then we can obtain $m$ models fine-tuned on $m$ tasks with parameters $\{\Theta_{1}, \Theta_{2}, ..., \Theta_{m}\}$. 
The fusion process in multi-task scenario, previously encapsulated by \autoref{eq:fusion_tailor}, is formulated as follows :
\begin{equation}
\begin{aligned}
 {\Theta}_\text{fusion} = \mathcal{F}( & {{\Theta}}_\text{sft}, {{\Theta}}_\text{pre}) \\
= \boldsymbol{M}_\text{agg} \odot ({{\Theta}}_\text{sft} + \boldsymbol{C}_\text{agg}) & + (I - \boldsymbol{M}_\text{agg}) \odot {{\Theta}}_\text{pre},
\end{aligned}
\end{equation}
where $\boldsymbol{M}_\text{agg}$ denotes the aggregated mask resulting from the union of individual task-specific masks $\boldsymbol{M}_i$, and $\boldsymbol{C}_\text{agg}$ signifies the average compensatory values determined for each task, calculated as follows:
\begin{equation}
   \boldsymbol{M}_\text{agg} = \bigcup_{i=1}^m \boldsymbol{M}_i, \quad \boldsymbol{C}_\text{agg} = ( \sum_{i=1}^m \boldsymbol{C}_i ) / m.
\end{equation}
The resultant model embodies a harmonized set of parameter adjustments and knowledge across multiple target tasks.
By aggregating the adjustments and selections from various tasks, this method not only underscores the model's enhanced capacity for generalization but also optimizes its performance across a diverse array of objectives.

\section{Proof}
\label{app:proof}
\subsection{Proof of \autoref{theorem:compensate}}
\label{app:proof_compensate}
\begin{theorem} (\autoref{theorem:compensate})
Given the alteration of a parameter from its fine-tuned state, $\theta_m^{\text{ft}}$, to its pre-trained state, $\theta_m^{\text{pre}}$, the optimal perturbation $\Delta \Theta$ for each retained fine-tuned parameters is calculated as follows:
\begin{equation}
\Delta \Theta^* = -\frac{\theta^m_{\text{sft}} - \theta^m_{\text{pre}}}{\left[\mathbf{H}^{-1}\right]_{mm}} \cdot \mathbf{H}_{:,m}^{-1},
\end{equation}
where $\mathbf{H}^{-1}$ denotes the inverse of the Hessian matrix and $\mathbf{H}_{:,m}^{-1}$ signifies the  $m$-th column of $\mathbf{H}^{-1}$. 
\end{theorem}

\textit{Proof.}
Based on \autoref{eq:layer_obj_true}, denote that $\mathcal{L}_\mathcal{T} = \left\|{\boldsymbol{\Theta}}^{\ell}_\text{sft} \mathbf{X}_{\mathcal{T}}^{\ell}-{\boldsymbol{\Theta}}^{\ell}_\text{fusion} \mathbf{X}_{\mathcal{T}}^{\ell}\right\|_2^2$. For clarity in the following proof, we will denote $\boldsymbol{\Theta}_\text{fusion}^\ell$ simply as $\boldsymbol{\Theta}$.
 We begin with a Taylor expansion of the loss function $\mathcal{L}_\mathcal{T}$ around the corresponding optimal parameters $\boldsymbol{\Theta}^*$. This expansion allows us to estimate the change in loss as we adjust the parameters from their optimal values. 
\begin{equation}
\mathcal{L}_\mathcal{T}(\boldsymbol{\Theta})=\mathcal{L}_\mathcal{T}\left({\boldsymbol{\Theta}^*}\right)+\mathcal{L}_\mathcal{T}^{\prime}\left(\boldsymbol{\Theta}^*\right)\left(\boldsymbol{\Theta}-\boldsymbol{\Theta}^*\right)+\frac{\mathcal{L}_\mathcal{T}^{\prime \prime}\left(\boldsymbol{\Theta}^*\right)}{2 !}\left(\boldsymbol{\Theta}-\boldsymbol{\Theta}^*\right)^2+\frac{\mathcal{L}_\mathcal{T}^{\prime \prime \prime}\left(\boldsymbol{\Theta}^*\right)}{3 !}\left(\boldsymbol{\Theta}-\boldsymbol{\Theta}^*\right)^3+\ldots.
\end{equation}

The goal is to minimize the impact on loss when a parameter is altered. The following equation represents the change in loss $ \Delta \mathcal{L}_\mathcal{T} $ when the parameters $ \boldsymbol{\Theta} $ deviate from their optimal setting $ \boldsymbol{\Theta}^* $.
\begin{equation}
\Delta \mathcal{L}_\mathcal{T}=\mathcal{L}_\mathcal{T}(\boldsymbol{\Theta})-\mathcal{L}_\mathcal{T}\left({\boldsymbol{\Theta}^*}\right)=\mathcal{L}_\mathcal{T}^{\prime}\left(\boldsymbol{\Theta}^*\right)\left(\boldsymbol{\Theta}-\boldsymbol{\Theta}^*\right) + \frac{\mathcal{L}_\mathcal{T}^{\prime \prime}\left(\boldsymbol{\Theta}^*\right)}{2 !}\left(\boldsymbol{\Theta}-\boldsymbol{\Theta}^*\right)^2+\frac{\mathcal{L}_\mathcal{T}^{\prime \prime \prime}\left(\boldsymbol{\Theta}^*\right)}{3 !}\left(\boldsymbol{\Theta}-\boldsymbol{\Theta}^*\right)^3+\ldots.
\end{equation}

To further refine our approach, we consider only the second-order term of the Taylor expansion, which involves the Hessian matrix $ \mathbf{H} $. This term is the primary contributor to the change in loss when small perturbations are made to the parameters.
\begin{equation}
    \Delta \mathcal{L}_\mathcal{T}=\left(\frac{\partial \mathcal{L}_\mathcal{T}}{\partial {\boldsymbol{\Theta}}}\right)^\top \cdot \Delta \boldsymbol{\Theta}+\frac{1}{2} \Delta \boldsymbol{\Theta}^\top \cdot \mathbf{H} \cdot \Delta \boldsymbol{\Theta}+O\left(\|\Delta \boldsymbol{\Theta}\|^3\right),
\end{equation}
where $ \Delta \boldsymbol{\Theta} =  \boldsymbol{\Theta} - \boldsymbol{\Theta}^*$.

In the layer-wise objective as delineated in \autoref{eq:layer_obj_true}, the constraint mandates that the cumulative absolute difference between the fusion parameters and their pre-trained counterparts, ${\boldsymbol{\Theta}}^{\ell}_\text{pre}$, must remain below a predefined threshold $\eta$. This restriction is crucial for retaining as much of the pre-trained knowledge as possible, thereby mitigating the risks of catastrophic forgetting. To effectively operationalize this constraint at an individual parameter level, we reformulate it into a more explicit condition, as shown in \autoref{eq:constriant}:
\begin{equation}
     \mathbf{e}_m^\top \cdot \Delta \boldsymbol{\Theta}+ \theta^m_\text{sft} = {\theta}_\text{pre}^m.
     \label{eq:constriant}
\end{equation}

This reformulation stipulates that for each fine-tuned parameter $\theta^m_\text{sft}$, the corresponding adjustment $\Delta \boldsymbol{\Theta}$ should result in a fusion parameter that equals the pre-trained parameter value ${\theta}_\text{pre}^m$. This constraint ensures that the knowledge embedded in each parameter of the pre-trained model is preserved to the maximum extent possible within the defined bounds of $\eta$.

To find the optimal perturbation $ \Delta \boldsymbol{\Theta} $ that minimizes the quadratic term in the loss change while satisfying the constraint, we set up the following minimization problem.
\begin{equation}
\min _{\Delta \boldsymbol{\Theta}}\left(\frac{1}{2} \Delta \boldsymbol{\Theta}^{\top} \mathbf{H} \Delta \boldsymbol{\Theta}\right), \quad \text { s.t. } \quad \mathbf{e}_m^{\top} \Delta \boldsymbol{\Theta}+\theta^m_\text{sft}=\theta_\text{pre}^m.
\end{equation}

In order to impose the best choice for the parameter to be removed, we can further consider the following constrained minimization problem.
\begin{equation}
     \min _m\left\{\left.\min _{\Delta \boldsymbol{\Theta}}\left(\frac{1}{2} \Delta \boldsymbol{\Theta}^\top \cdot \mathbf{H} \cdot \Delta \boldsymbol{\Theta}\right) \right\rvert\, \mathbf{e}_m^\top \cdot \Delta \boldsymbol{\Theta}+\theta^m_\text{sft}={\theta}_\text{pre}^m \right\}.
     \label{eq:last_objective}
\end{equation}

We use the method of Lagrange multipliers to solve this constrained optimization problem. The Lagrangian $ L $ integrates the objective function with the constraint, incorporating the Lagrange multiplier $ \lambda $.
\begin{equation}
    L(\Delta \boldsymbol{\Theta})=\frac{1}{2} \Delta \boldsymbol{\Theta}^\top \cdot \mathbf{H} \cdot \Delta \boldsymbol{\Theta}+\lambda\left(\mathbf{e}_m^\top \cdot \Delta \boldsymbol{\Theta}+ \theta^m_\text{sft}-{\theta}_\text{pre}^m \right).
\end{equation}

Taking the derivative of the Lagrangian with respect to $ \Delta \boldsymbol{\Theta} $ and setting it to zero gives us the direction in which the Lagrangian is stationary, which is necessary for finding the optimal point.
\begin{equation}
\frac{\partial  L(\Delta \boldsymbol{\Theta})}{\partial \Delta \boldsymbol{\Theta} } = -\mathbf{X}^\top\mathbf{X} \Delta\boldsymbol{\Theta} + \lambda \mathbf{e}_m^\top = 0.
\end{equation}


Furthermore, the derivative of the Lagrangian with respect to the Lagrange multiplier $\lambda$ must also equal zero to satisfy the constraint. This derivative yields:
\begin{equation}
\frac{\partial L(\Delta \boldsymbol{\Theta})}{\partial \lambda} = \mathbf{e}_m^\top \cdot \Delta \boldsymbol{\Theta}+ \theta^m_\text{sft}-{\theta}_\text{pre}^m = 0.
\end{equation}

By solving these equations, we obtain the expression for $\Delta \boldsymbol{\theta}$, which minimizes the objective function while satisfying the constraint that the final weights should align with the pre-update weights to avoid catastrophic forgetting:
\begin{equation}
\Delta \boldsymbol{\Theta} = (\mathbf{X}^\top\mathbf{X})^{-1}\lambda  \mathbf{e}_m^\top.
\end{equation}


The Lagrange multiplier $\lambda$ is then computed based on the imposed constraint and the inverse of the Hessian matrix, leading to the final update rule for the weight perturbation:
\begin{equation}
\lambda = ( \theta^m_\text{sft}-{\theta}_\text{pre}^m) \cdot \frac{1}{(\mathbf{H}^{-1})_{mm}}.
\end{equation}

The corresponding optimal perturbation so obtained is as follows when altering the $m$-th parameter $\theta_m$:
\begin{equation}
\Delta \boldsymbol{\Theta}^*_m = -\frac{ \theta^m_\text{sft}-{\theta}_\text{pre}^m}{\left[\mathbf{H}^{-1}\right]_{ii}} \cdot \mathbf{H}^{-1} \cdot \mathbf{e}^m = -\frac{ \theta^m_\text{sft}-{\theta}_\text{pre}^m}{\left[\mathbf{H}^{-1}\right]_{ii}} \cdot \mathbf{H}^{-1}_{:,m}.
\label{eq:delta}
\end{equation}

The proof of \autoref{theorem:compensate} is finished.

\qed

\subsection{Proof of \autoref{theorem:sensitive}}
\label{app:proof_sensitive}

\begin{theorem} (\autoref{theorem:sensitive})
Consider a layer $\ell$ within an MLLM $\mathcal{M}$, and let $\theta_m$ represent a parameter at index $m$. Altering $\theta_m$ from its fine-tuned state $\theta^m_{\text{sft}}$ to its pre-trained state $\theta^m_{\text{pre}}$, induces a increase $\Delta \mathcal{L}_\mathcal{T}$ in the model's loss for a task $\mathcal{T}$, quantified as:
\begin{equation}
\Delta \mathcal{L}_\mathcal{T} =\frac{(\theta^m_{\text{sft}} - \theta^m_{\text{pre}})^2}{\left[\mathbf{H}^{-1}\right]_\text{mm}},
\end{equation}
where $\mathbf{H}^{-1}$ denotes the inverse of the Hessian matrix and $\left[\mathbf{H}^{-1}\right]_{mm}$ is its $m$-th diagonal element. The Hessian matrix $\mathbf{H}=\nabla_{\theta^m_\text{sft}}^2 \mathcal{L}_{\mathcal{T}}$ takes into account the local geometry of the loss at a given point $\theta^m_\text{sft}$ assuming $\mathcal{L}_\mathcal{T}$ is second-order differentiable.
\end{theorem}

\textbf{Proof.} The change in loss due to the optimal perturbation obtained in \autoref{eq:delta} that aligns with the pre-trained model's parameters is quantified, giving us a measure of how the loss increases as we modify the network.
\begin{equation}
\begin{aligned}
   \Delta \mathcal{L}_\mathcal{T} &=\left(\frac{\partial \mathcal{L}_\mathcal{T}}{\partial {\boldsymbol{\Theta}}}\right)^\top \cdot \Delta \boldsymbol{\Theta}+\frac{1}{2} \Delta \boldsymbol{\Theta}^\top \cdot \mathbf{H} \cdot \Delta \boldsymbol{\Theta}+O\left(\|\Delta \boldsymbol{\Theta}\|^3\right) \\
   &  \approx \frac{1}{2} \Delta \boldsymbol{\Theta}^\top \cdot \mathbf{H} \cdot \Delta \boldsymbol{\Theta} \\
   &  \approx \frac{( \theta^m_\text{sft}-{\theta}_\text{pre}^m)^2}{2\left[\mathbf{H}^{-1}\right]_{mm}}.
\end{aligned}
\end{equation}

Returning to the challenge presented in \autoref{eq:last_objective}, the optimal selection of the parameter index $m$ involves identifying and removing the parameter $\theta_m$ that results in the smallest increase in loss. This increase in loss is quantified as the sensitivity-based statistic $s_\varepsilon$, as defined in \autoref{eq:s_var}:
\begin{equation}
    s_\varepsilon = \frac{( \theta^m_\text{sft}-{\theta}_\text{pre}^m)^2}{2\left[\mathbf{H}^{-1}\right]_{mm}}.
    \label{eq:s_var}
\end{equation}
We calculate this statistic for all parameters, then rank them in descending order based on the magnitude of $s_\varepsilon$. 

The proof of \autoref{theorem:sensitive} is finished.

\qed

\section{More Details about Experiments}
\label{app:experments}

\subsection{Implementation Details} 
\label{app:experments_computation}

\textbf{Standard Fine-tuning.} 
In the case of InstructBLIP, our fine-tuning procedure adhered to the official codebase~\footnote{https://github.com/salesforce/LAVIS} guidelines with a batch size of 12 for each task, a maximum of 5 epochs, and a learning rate of 1e-5. Similarly, for LLaVA, we followed the official fine-tuning protocols~\footnote{https://github.com/haotian-liu/LLaVA}, setting the maximum epoch to 1. The initial learning rate was configured at 2e-4 for fine-tuning on Flickr30k and 1e-4 for GQA, with a batch size of 64. All experiments were conducted on 8 NVIDIA V100 GPU  with  32GB of memory.

\textbf{Computation Complexity of Model Tailor.} In the implementation of Model Tailor, we have employed an efficient computational technique called SparseGPT~\cite{frantar2023sparsegpt} to calculate the inverse Hessian matrix, a critical component in the OBS calculation. The computational complexity of calculating the inverse Hessian as described in SparseGPT involves three main components:
(1) Initial Hessian Computation: The initial Hessian matrix is computed with a time complexity of $\mathcal{O}(n · d_\text{col}^2)$, where $n$ is the number of input samples and $d_\text{col}$ is the dimensionality of the matrix column. By choosing $n$ to be a small multiple of $d_\text{col}$, we achieve stable results while maintaining efficiency.
(2) Hessian Inversion Iteration: Iterating through the inverse Hessian requires $\mathcal{O}(d_\text{col}^3)$ time complexity, which is manageable even for larger models.
(3) Reconstruction Process: The pruning or reconstruction process, based on the inverse Hessian, is bounded by the complexity of $\mathcal{O}(d_\text{col}^3 + d_\text{row}^2 d_\text{col})$, where $d_row$ represents the dimensionality of the matrix row. This ensures that even for models with extensive parameters, the process remains computationally feasible.
In summary, the overall computational complexity aligns with $\mathcal{O}(d_\text{hidden}^3)$, considering a Transformer model's hidden dimensionality $d_\text{hidden}$. This demonstrates a substantial efficiency improvement over exact reconstruction methods, confirming that our approach is computationally practical even for very large models.

\textbf{Combination with LoRA.} We adhered to the standard LoRA fine-tuning paradigm provided by the official code repository\footnote{https://github.com/haotian-liu/LLaVA} of LLaVA. For fine-tuning on Flickr30k, we set the initial learning rate at 2e-4, and for GQA, the initial learning rate was set at 1e-4. All other hyperparameter settings remained consistent with the original setup of LLaVA.
After the fine-tuning process on LLaVA-1.5 using LoRA, we integrate the incremental updates from LoRA throughout the entire fine-tuned model. This means that our Model Tailor method is applied across the full model, encompassing both the MLP mapper and the entirety of the LLM, which totals approximately 6.7 billion parameters. Therefore, to manage the volume of selected parameters effectively and ensure it remains substantial but not excessive, we employed a sparsity level of 5\%, which corresponds to approximately 335 million parameters. 
 This approach ensures that Model Tailor works in conjunction with the enhancements provided by LoRA, facilitating a comprehensive adaptation across the model's expansive parameter space.


\subsection{Evaluation Metrics}
\label{app:experments_metric}
To rigorously assess the effectiveness of our method in mitigating catastrophic forgetting on MLLMs, we utilize two key evaluation metrics: the Average Performance and the H-score. The H-score, in particular, is a novel metric designed to provide a balanced evaluation between original and target tasks. It is defined as the harmonic mean of the average performance on the original tasks, $\operatorname{Avg}(P_\text{origin})$, and the average performance on the target tasks, $\operatorname{Avg}(P_\text{target})$. The formula for the H-score is as follows:
\begin{equation}
P_H = \frac{2 \times \operatorname{Avg}(P_\text{origin}) \times \operatorname{Avg}(P_{target})}{\operatorname{Avg}(P_\text{origin}) + \operatorname{Avg}(P_\text{target})}.
\end{equation}
The rationale behind the introduction of the H-score is to circumvent the potential overemphasis on the performance of original tasks, particularly as their number increases.

\subsection{More Results}
\label{app:experments_results}
\textbf{Quantitative Results on Multi-Task Setting.}
\autoref{tab:multitask_instructblip} and \autoref{tab:multitask_llava} present the quantitative outcomes of Model Tailor within a multi-task context. The data delineates a clear enhancement in performance on pre-trained tasks when integrating masks and compensatory values from two distinct tasks. This amalgamation of knowledge fosters a generalized proficiency, underscoring the method's improved adaptability and robustness across varied tasks.

\textbf{Quantitative results of Model Tailor combined with LoRA.}
\autoref{tab:lora_llava} details the quantitative performance of Model Tailor when amalgamated with LoRA on the LLaVA-1.5 model. The results demonstrate remarkable performance boosts, regardless of whether the model was fine-tuned on Flickr30k or OKVQA. This indicates the potentiated benefits of combining Model Tailor's approach with LoRA's fine-tuning capabilities.

\begin{figure*}[t]
  \centering
    \includegraphics[width=6.8in]{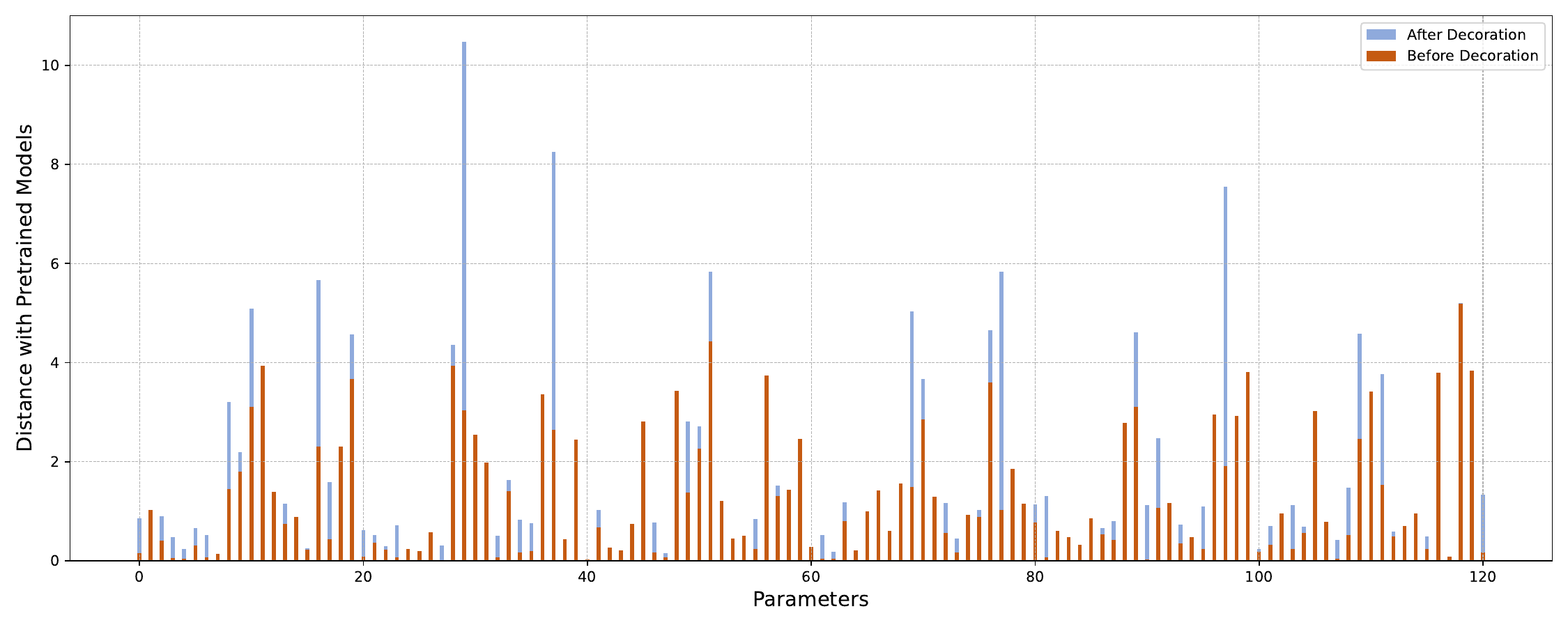}
  \caption{\textbf{Visualization of Decorator's Effectiveness.} This bar graph contrasts the parameter distances from their pre-trained counterparts before and after the application of the patch decorator. The decorator moves parameters closer to their fine-tuned states, demonstrates the decorator's capacity to counterbalance the exclusion of certain parameters.
  }
  \label{fig:decorator}
\end{figure*}

\textbf{Visualization of decorator.}
\autoref{fig:decorator} visualizes the parametric shifts pre and post compensation. The vertical axis quantifies the absolute difference in cumulative sums between each parameter and its corresponding pre-trained value. In essence, taller bars within the histogram signify a greater deviation towards the fine-tuned model's parameters. The illustration evidences the decorator's role in nudging the model parameters closer to the fine-tuned state, thereby compensating for the performance impact on the target task due to the removal of certain parameters (while maintaining pre-trained values).


\begin{table*}[t]
  \caption{
  \textbf{H-score and average performance on InstructBLIP (Vicuna-7B) with a sparsity ratio $\rho=10\%$ under multi-task setting.} The optimal and sub-optimal results are denoted respectively by boldface and underlining for emphasis.
  }
  \label{tab:multitask_instructblip}
  \centering
  \resizebox{1\linewidth}{!}{
  \begin{tabular}{l|ccccccc|cc|>{\columncolor{tabhighlight}}c >{\columncolor{tabhighlight}}c}

      \bottomrule

      \toprule
      \multirow{2}{*}{Method} &  \multicolumn{7}{c}{\textbf{Pretrain tasks}} & \multicolumn{2}{c}{\textbf{Target task}}\\
      & 
      \multicolumn{1}{c}{\textcolor{someblue}{COCO}} &
      \multicolumn{1}{c}{\textcolor{someblue}{NoCaps-in}} &
      \multicolumn{1}{c}{\textcolor{someblue}{NoCaps-near}} &
      \multicolumn{1}{c}{\textcolor{someblue}{NoCaps-out}} &
      \multicolumn{1}{c}{\textcolor{someblue}{OKVQA}} &
      \multicolumn{1}{c}{\textcolor{someblue}{AOKVQA}} &
      \multicolumn{1}{c}{\textcolor{someblue}{VQAv2}} &
         \multicolumn{1}{c}{\textcolor[rgb]{0.773,0.353,0.067}{GQA}} &
      \multicolumn{1}{c}{\textcolor[rgb]{0.773,0.353,0.067}{Flickr30k}} &
      \multicolumn{1}{c}{Avg} &
      \multicolumn{1}{c}{Hscore} 
      \\  
      
      \midrule
      {Flickr30k-Fintuned}   & 139.8 & 115.8 & \underline{124.3}&\underline{123.4} & \textbf{57.63}  & 57.84  &76.19& 48.58 & \textbf{94.4} & \underline{93.10} &\textbf{83.12}      \\
  {GQA-Fintuned}  & \underline{141.8} & \textbf{116.5} & 121.8& 120.7& 56.15  & \underline{58.33}  &\underline{76.28}& \textbf{53.98} &82.7 &92.03 &80.79           \\
  {Hybrid-Flickr30k-GQA}   & \textbf{141.9} & \underline{116.3} & \textbf{125.4}& \textbf{125.5}& \underline{57.09}  & \textbf{58.64}   &\textbf{76.69} & \underline{52.00} & \underline{90.0} &\textbf{93.72}  &\textbf{83.12}          \\
  \bottomrule

  \toprule

  \end{tabular}
  }
  \end{table*}

\begin{table*}[t]
  \caption{
  \textbf{H-score and average performance on LLaVA-1.5 (Vicuna-7B) with a sparsity ratio $\rho=10\%$ under multi-task setting.} The optimal and sub-optimal results are denoted respectively by boldface and underlining for emphasis.
  }
  \label{tab:multitask_llava}
  \centering
  \resizebox{1\linewidth}{!}{
  \begin{tabular}{l|cccccccc|cc| >{\columncolor{tabhighlight}}c >{\columncolor{tabhighlight}}c}

      \bottomrule

      \toprule
      \multirow{2}{*}{Method} &  \multicolumn{8}{c}{\textbf{Pretrain tasks}} & \multicolumn{2}{c}{\textbf{Target task}}\\
      & 
      \multicolumn{1}{c}{\textcolor{someblue}{VQAv2}} &
      \multicolumn{1}{c}{\textcolor{someblue}{GQA}} &
      \multicolumn{1}{c}{\textcolor{someblue}{VizWiz}} &
      \multicolumn{1}{c}{\textcolor{someblue}{SQA}} &
      \multicolumn{1}{c}{\textcolor{someblue}{TextVQA}} &
      \multicolumn{1}{c}{\textcolor{someblue}{POPE}} &
      \multicolumn{1}{c}{\textcolor{someblue}{MM-Bench}} &
       \multicolumn{1}{c}{\textcolor{someblue}{MM-Bench-CN}} &
        \multicolumn{1}{c}{\textcolor{someorange}{OKVQA}} &
      \multicolumn{1}{c}{\textcolor{someorange}{Flickr30k}} &
      \multicolumn{1}{c}{Avg} &
      \multicolumn{1}{c}{Hscore} 
      \\  
      
      \midrule
      {Flockr30k-fine-tuned}  &73.21 &52.49 &42.28 &{67.15} &43.89 &{82.88}& \underline{63.40} & \underline{56.15} & 7.05 & \textbf{75.4} & 62.72 & \underline{64.80}        \\
  {OKVQA-fine-tuned} & \underline{76.25} & \underline{60.39}& \textbf{46.49}&\underline{69.51} &\underline{54.88} &\underline{85.44} & {63.32}& 54.21& \textbf{38.1} & 0.01 & \underline{65.18} & 54.01   \\
  {Hybrid-Flickr30k-OKVQA}  & \textbf{76.88} & \textbf{61.29}  &\underline{45.88} &  \textbf{72.41}& 53.76 &\textbf{85.96}  &\textbf{63.77} & \textbf{57.72} &\underline{30.2} & \underline{59.7}  & \textbf{65.71} & \textbf{69.52}     \\
  \bottomrule

  \toprule

  \end{tabular}
  }
  \end{table*}

  \begin{table*}[t]
  \caption{\textbf{Combination with LoRA on LLaVa1.5 (Vicuna-7b) with a sparsity ratio $\rho = 5\%$.} The optimal and sub-optimal results are denoted respectively by boldface and underlining for emphasis.}
  \label{tab:lora_llava}
  \centering
  \resizebox{1\linewidth}{!}{
  \begin{tabular}{l|cccccccc|c| >{\columncolor{tabhighlight}}c >{\columncolor{tabhighlight}}c}

      \bottomrule
      
      \toprule
      \multirow{2}{*}{Method} &  \multicolumn{8}{c}{\textbf{Pre-trained tasks}} & \multicolumn{1}{c}{\textbf{Target task}}\\
      & 
      \multicolumn{1}{c}{\textcolor{someblue}{VQAv2}} &
      \multicolumn{1}{c}{\textcolor{someblue}{GQA}} &
      \multicolumn{1}{c}{\textcolor{someblue}{VizWiz}} &
      \multicolumn{1}{c}{\textcolor{someblue}{SQA}} &
      \multicolumn{1}{c}{\textcolor{someblue}{TextVQA}} &
      \multicolumn{1}{c}{\textcolor{someblue}{POPE}} &
      \multicolumn{1}{c}{\textcolor{someblue}{MM-Bench}} &
       \multicolumn{1}{c}{\textcolor{someblue}{MM-Bench-CN}} &
      \multicolumn{1}{c}{\textcolor[rgb]{0.773,0.353,0.067}{Flickr30k}} &
      \multicolumn{1}{c}{Avg} &
      \multicolumn{1}{c}{Hscore} 
      \\  
      
      \midrule
  {LoRA} &  76.35 & 58.64&50.25 &65.13 & 55.18 & \textbf{84.91}& 64.43&53.09 & 62.5 & 63.39 &62.99       \\
  {Ours + LoRA}  &  \textbf{78.59} & \textbf{61.39} & \textbf{53.2}  & \textbf{69.61} & \textbf{58.01} & 84.22  & \textbf{65.03}& \textbf{57.82} &81.4  & \textbf{67.70} & \textbf{72.89}       \\
  \bottomrule

  \toprule

   \multirow{2}{*}{Method}&  \multicolumn{8}{c}{\textbf{Pre-trained tasks}} & \multicolumn{1}{c}{\textbf{Target task}}\\
      & 
        \multicolumn{1}{c}{\textcolor{someblue}{VQAv2}} &
      \multicolumn{1}{c}{\textcolor{someblue}{GQA}} &
      \multicolumn{1}{c}{\textcolor{someblue}{VizWiz}} &
      \multicolumn{1}{c}{\textcolor{someblue}{SQA}} &
      \multicolumn{1}{c}{\textcolor{someblue}{TextVQA}} &
      \multicolumn{1}{c}{\textcolor{someblue}{POPE}} &
      \multicolumn{1}{c}{\textcolor{someblue}{MM-Bench}} &
       \multicolumn{1}{c}{\textcolor{someblue}{MM-Bench-CN}} &
      \multicolumn{1}{c}{\textcolor[rgb]{0.773,0.353,0.067}{OKVQA}} &
      \multicolumn{1}{c}{Avg} &
      \multicolumn{1}{c}{Hscore} 
      \\  
      
      \midrule
  {LoRA}  &71.21 & 52.87&37.09 &28.15 &54.36 & 67.81& 35.91 &29.47 &59.25 &48.45  &52.48      \\
  {Ours + LoRA}  & \textbf{72.44}  &  \textbf{54.97} & \textbf{40.09} & \textbf{57.56} &  \textbf{55.73} & 67.73 & \textbf{56.01} & \textbf{46.39} & \textbf{59.82} &\textbf{56.75} & \textbf{58.04}  \\
  \bottomrule

  \toprule

  \end{tabular}
  }
  \end{table*}

\end{document}